\documentclass[letterpaper]{article} 
\usepackage{aaai2026}
\usepackage{times}  
\usepackage{helvet}  
\usepackage{courier}  
\usepackage[hyphens]{url}  
\usepackage{graphicx} 
\usepackage{natbib}  
\usepackage{caption} 
\usepackage{algorithm}
\usepackage{algorithmic}

\usepackage{subcaption} 
\usepackage{amsfonts}
\usepackage{amsthm}
\usepackage{amsmath}

\usepackage{textcomp}
\usepackage{bm}
\usepackage{makecell}
\usepackage{multirow}
\usepackage{enumitem}

\usepackage{newfloat}
\usepackage{listings}
\usepackage{colortbl}
\usepackage{multicol}
\usepackage{xspace}
\usepackage{bbding}
\usepackage{pifont}
\usepackage{tabularx}
\usepackage{booktabs}
\usepackage{array}    
\usepackage{tcolorbox}

\newtheorem{theorem}{{Theorem}}

\newtheorem{definition}{{Definition}}

\newtheorem{problem}{{Problem}}

\DeclareCaptionStyle{ruled}{labelfont=normalfont,labelsep=colon,strut=off} 
\floatstyle{ruled}
\newfloat{listing}{tb}{lst}{}
\floatname{listing}{Listing}
\title{Unlocking Multi-Modal Potentials for Link Prediction on\\ Dynamic Text-Attributed Graphs}

\author{
Yuanyuan Xu\textsuperscript{\rm 1}, Wenjie Zhang\textsuperscript{\rm 1}, Ying Zhang\textsuperscript{\rm 2}, Xuemin Lin\textsuperscript{\rm 3}, Xiwei Xu\textsuperscript{\rm 4}
}
\affiliations{
    \textsuperscript{\rm 1}University of New South Wales, \textsuperscript{\rm 2}Zhejiang Gongshang University, \textsuperscript{\rm 3}Shanghai Jiao Tong University, \textsuperscript{\rm 4}CSIRO Data61\\
}

\usepackage{bibentry}

\begin{document}

\maketitle

\begin{abstract}
Dynamic Text-Attributed Graphs (DyTAGs) are a novel graph paradigm that captures evolving temporal events (edges) alongside rich textual attributes. Existing studies can be broadly categorized into TGNN-driven and LLM-driven approaches, both of which encode textual attributes and temporal structures for DyTAG representation. We observe that DyTAGs inherently comprise three distinct modalities: temporal, textual, and structural, often exhibiting completely disjoint distributions. However, the first two modalities are largely overlooked by existing studies, leading to suboptimal performance. To address this, we propose MoMent, a multi-modal model that explicitly models, integrates, and aligns each modality to learn node representations for link prediction. Given the disjoint nature of the original modality distributions, we first construct modality-specific features and encode them using individual encoders to capture correlations across temporal patterns, semantic context, and local structures. Each encoder generates modality-specific tokens, which are then fused into comprehensive node representations with a theoretical guarantee. To avoid disjoint subspaces of these heterogeneous modalities, we propose a dual-domain alignment loss that first aligns their distributions globally and then fine-tunes coherence at the instance level. This enhances coherent representations from temporal, textual, and structural views. Extensive experiments across seven datasets show that MoMent achieves up to $17.28\%$ accuracy improvement and up to $31\times$ speed-up against eight baselines.
\end{abstract}

\section{Introduction}
Dynamic Text-Attributed Graphs (DyTAGs) are a recent development to represent systems with evolving structures and text attributes over time, where nodes and edges are enriched with text attributes. DyTAGs are critical for modeling complex real systems, such as e-commerce platforms~\cite{zhao2023time,kazemi2020representation,10.1145/3580305.3599341,10.1145/3637528.3671962} and social networks~\cite{pareja2020evolvegcn,sun2022ddgcn,luo2023hope,song2019session}. For instance, in social networks, nodes represent posts with text attributes, and edges may be annotated with user comments, allowing users to engage in behaviors at any time. To represent DyTAGs and support real applicability, there is a pioneering exploration DTGB~\cite{zhang2024dtgb} that leverages Pre-trained Language Models (PLMs) to encode text attributes and then integrates them with Temporal Graph Neural Networks (TGNNs)~\cite{kumar2019predicting,trivedi2019dyrep,xu2020inductive,tgn,wang2021inductive,congwe,yu2023towards} for node representations. Subsequently, LKD4DyTAG~\cite{roy2025llm} employs a knowledge distillation framework to transfer knowledge from the text-based edge representation of a teacher Large Language Model (LLM) model to the spatio-temporal representation of a student Graph Neural Network (GNN) for DyTAG representation. These explorations predominantly adopt edge-centric modeling that emphasizes local connectivity (node external) but often overlooks node internal semantic dynamics (temporal/textual), thereby compromising model performance.


\begin{figure*}
    \centering
    \subfloat[A dynamic text-attributed graph]{\includegraphics[width=0.39\linewidth]{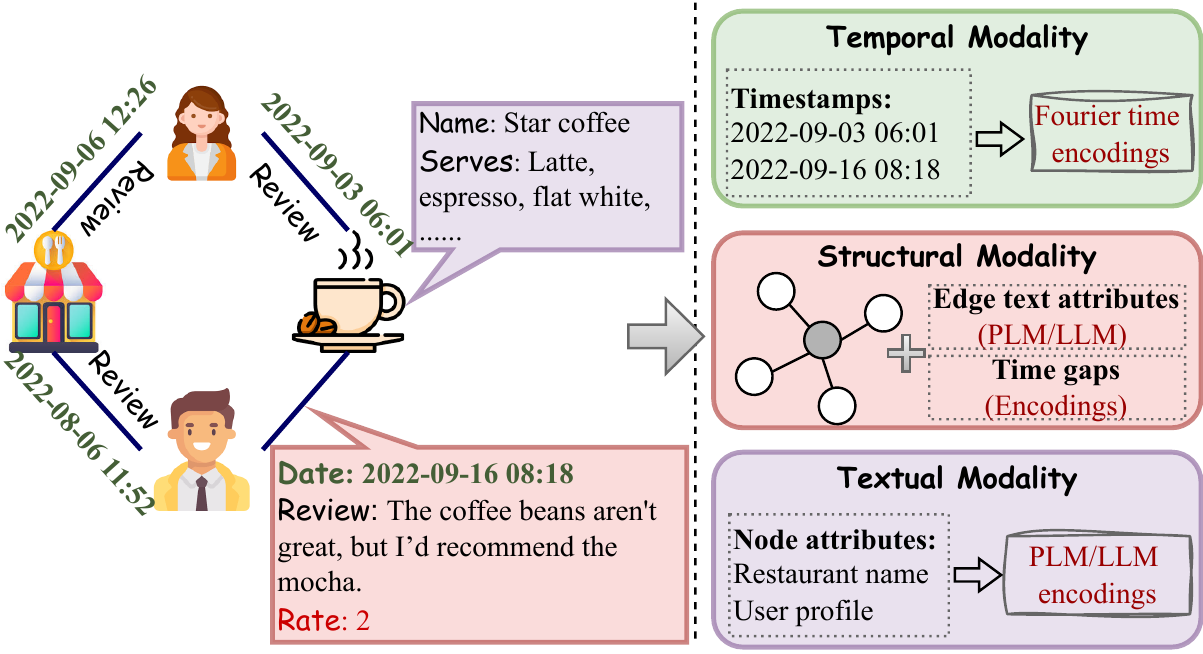}\label{fig:1a}}\;
    \subfloat[ KDE distribution of modalities]{
            \includegraphics[width=0.25\linewidth]{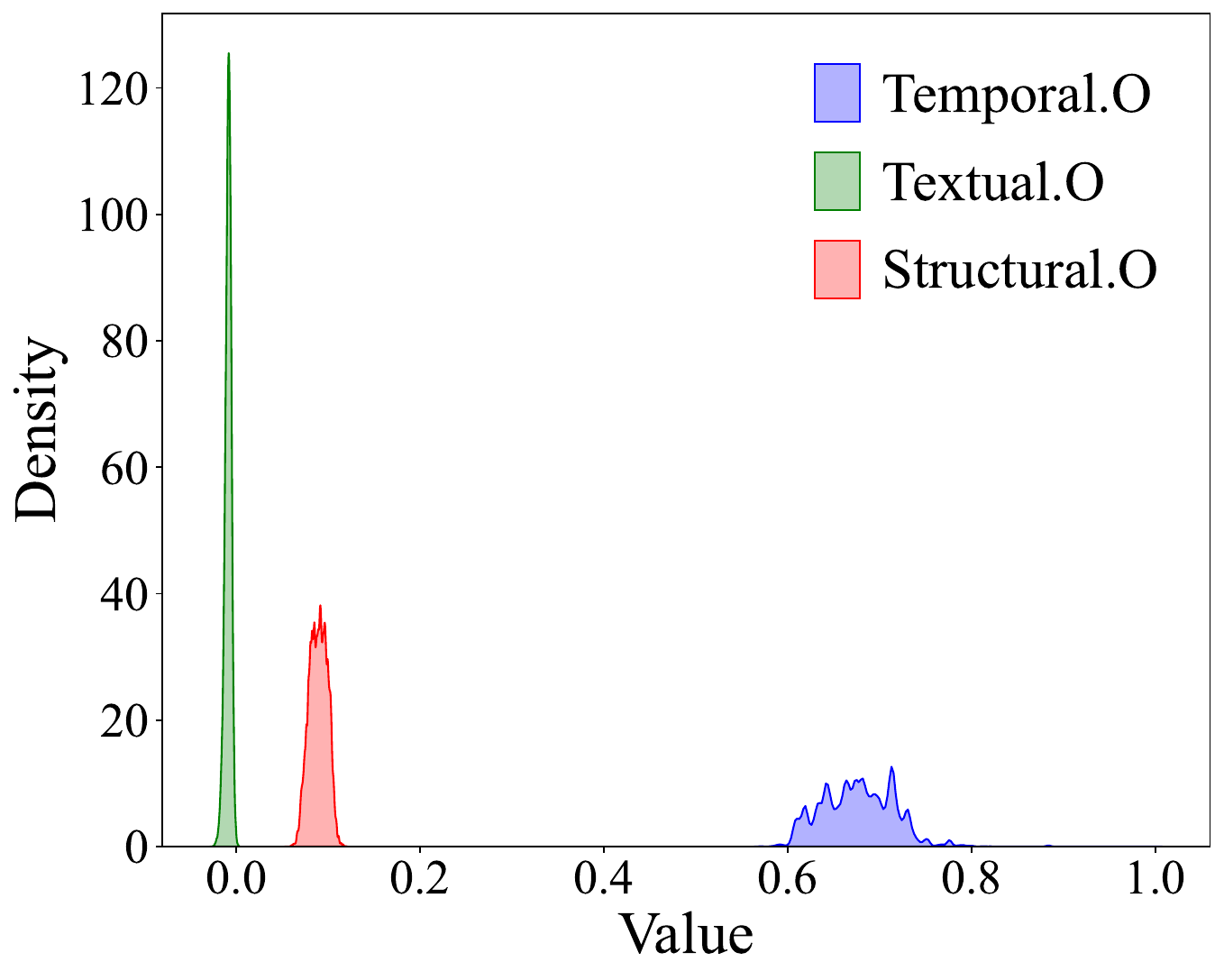}
        \label{fig:1b}}  \;
    \subfloat[Text attribute statistics]{
            \includegraphics[width=0.27\linewidth]{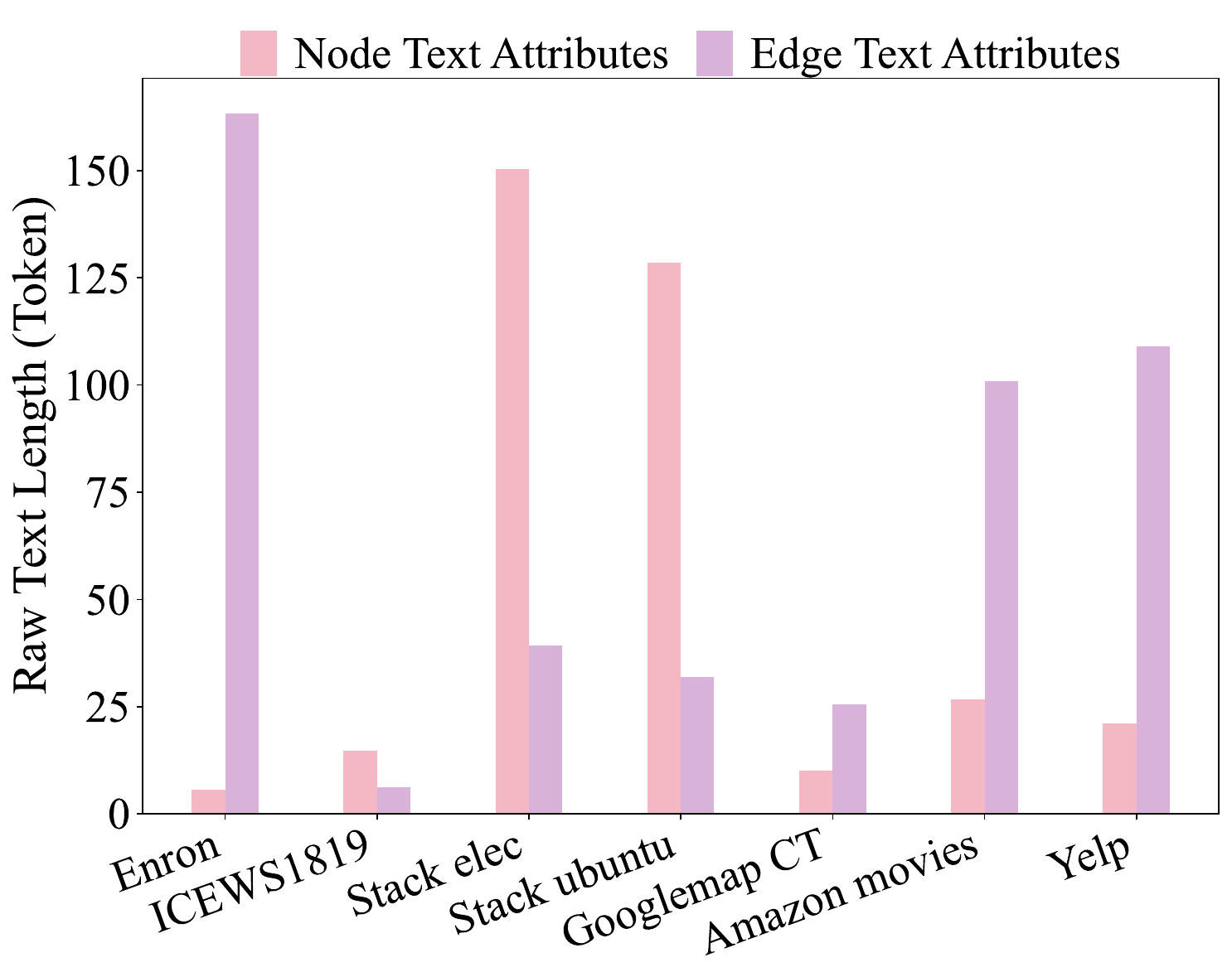}
        \label{fig:1c}}
    \caption{(1) A toy example of DyTAGs, together with its three modalities. (b) KDE distribution of three modalities in a real DyTAG (Enron), where `X.O' indicates original features. (c) The average text length of edges and nodes in $7$ DyTAGs.}
    \label{fig:1}
\end{figure*}

\noindent\textbf{Observation for multi-model nature of DyTAGs.} Here, we first analyze unique characteristics of DyTAGs, illustrated by the toy example in Fig.~\ref{fig:1a}, to establish a foundation for effective modeling. Taking a \textit{node-centric} perspective, we identify DyTAGs as inherently multi-modal, which distinguishes them from traditional dynamic graphs: (1) \underline{Structural Modality}: Similar to dynamic graphs, each node interacts with multiple neighbors through evolving edges, where \textit{edge text attributes and timestamps are inherently associated with edges}, encoding external semantic dynamics of nodes. (2) \underline{Temporal Modality}: Beyond structural changes, nodes initiate interactions at different timestamps, capturing temporal evolution patterns (\textit{e.g.}, frequency) that preserve node internal dynamics. (3) \underline{Textual Modality}: Unlike dynamic graphs, nodes in DyTAGs are inherently enriched with raw text attributes, introducing internal semantic contexts of nodes.

To better understand these three modalities, we analyze their Kernel Density Estimation (KDE) distributions using the Enron dataset, as shown in Fig.~\ref{fig:1b}, where the structural modality is derived from the $20$ neighbors of each node. Temporal modality has a dispersed distribution with values from $0.6$ to $0.8$, reflecting global variability in event timings, whereas textual modality is compact and concentrated near $0$, indicating high semantic regularity. The structural modality exhibits a moderate distribution, capturing structural diversity and local variability in edge attributes. Furthermore, the temporal, textual, and structural modalities exist in different value spaces with disjoint distributions, highlighting the necessity for multi-modal modeling and alignment while posing the following challenges.


\noindent\textbf{Challenge \uppercase\expandafter{\romannumeral1}: How to unlock the potential of each modality for comprehensive representation?}
Existing approaches~\cite{zhang2024dtgb,roy2025llm} typically entangle temporal and/or textual modalities with structure learning, confining attention to local neighbors. Specifically, timestamps are merely treated as auxiliary features during local structure learning via TGNNs, preventing these models from capturing rich temporal patterns such as frequency.
Subsequently, LKD4DyTAG~\cite{roy2025llm} leverages LLMs with prompt-based mechanisms to enhance edge textual semantics from an edge-centric view, working with GNNs for DyTAG representation. 
Yet, our statistics (Fig.~\ref{fig:1c}) show that the text attributes attached to nodes and edges in existing DyTAGs are relatively short, only $3$ to $164$ tokens, limiting the gains attainable from LLMs. Moreover, Fig.~\ref{fig:1c} shows that node text attributes convey even greater semantic richness, judged by their length, than edge attributes, a strength that existing approaches largely ignore. These observations lead to a key question: how can we design a lightweight framework that fully exploits temporal, textual, and structural modalities for DyTAG representation?

\noindent\textbf{Challenge \uppercase\expandafter{\romannumeral2}: How to align the heterogeneous modalities for coherent representation?} Fig.~\ref{fig:1b} shows that the three modalities in DyTAGs (\textit{i.e.}, temporal, textual, and structural) exhibit distinct value ranges and density distributions, reflecting their inherently different characteristics. This confirms that the modalities are largely disjoint in nature. While modeling each modality individually can capture its unique characteristics, combining them without proper alignment often leads to disjoint latent spaces. This misalignment results in modality inconsistency, making it difficult to generate coherent node representations and ultimately compromising model performance. Therefore, it is crucial to develop an alignment mechanism for coherent representation.


To address these challenges, we propose MoMent, a lightweight yet effective model that fully leverages each modality and effectively aligns those residing in different value spaces for link prediction on dynamic text-attributed graphs. Guided by the density patterns illustrated in Fig.~\ref{fig:1b}, MoMent employs a self-attention-only architecture with modality-specific encoders to capture temporal dynamics from recent node behavior timestamps, textual semantics from node attributes, and structural correlations from the local graph topology. Such a learning mechanism extracts both internal and external semantic dynamics, addressing \textbf{Challenge \uppercase\expandafter{\romannumeral1}}. The encoder outputs are fused to generate comprehensive node representations, where we theoretically demonstrate that our multi-modal modeling enhances information gain over a single structural modality. 
To tackle \textbf{Challenge \uppercase\expandafter{\romannumeral2}} and avoid the modality misalignment, MoMent first aligns temporal and textual distributions globally, then refines coherence among all three modalities at the instance level. 
This dual-domain alignment loss enforces coherence across temporal, textual, and structural modalities, preventing disjoint latent spaces and enabling the generation of comprehensive and coherent node representations.



Our contributions can be summarized as

\begin{itemize}
    \item We propose MoMent for modeling, aligning, and integrating the multi-modal nature of DyTAGs, which can capture internal and external semantic dynamics for node representations. We theoretically demonstrate that MoMent can achieve additional conditional information gain from internal semantic dynamics.

    \item MoMent investigates the self-attention-only architecture to capture temporal patterns, textual semantics, and structural correlations based on effective modality construction, while ensuring temporal-semantic-structural coherence via our dual-domain alignment loss.    
    
   \item Extensive experiments demonstrate that MoMent yields better prediction performance than eight baselines over seven datasets, with an average of $7.05\%$ improvement on inductive link prediction with minimal time costs. We visualize the subspace alignment of modality-specific tokens of MoMent.

\end{itemize}

\section{Related Work}
\subsection{Dynamic Text-Attributed Graph Learning}
The Dynamic Text-Attributed Graph (DyTAG) is a new concept, recording the evolving dynamic structures and text attributes simultaneously. Behind this new concept, there is a new and comprehensive benchmark (DTGB)~\cite{zhang2024dtgb} for DyTAG learning. Concretely, DTGB first introduced DyTAG datasets from multiple domains and standardized the evaluation process. Then, it presented an intuitive framework that learned raw text attributes of nodes and edges by pre-trained language models and then integrated them into existing TGNNs. Although text attributes are employed to learn node representations, DTGB with various TGNNs~\cite{kumar2019predicting,trivedi2019dyrep,xu2020inductive,congwe,yu2023towards} focuses exclusively on local learning based on the graph topology, where timestamps and edge/node attributes are merely treated as an optional supplement for structure learning, leading to the underexploration of temporal and textual modalities. LKD4DyTAG~\cite{roy2025llm} distilled LLM-based edge semantics into a GNN that models spatio-temporal structure, but it treated time information merely as an auxiliary for local structure learning. However, this fails to capture temporal patterns for node representations, like node behavior frequency, while edge text modelling based on LLMs would lead to substantial computation costs. 
In contrast, we propose a node-centric modeling that fully exploits three modalities to enhance DyTAG representation. 
Besides, we investigate the use of self-attention-only encoders, eliminating reliance on LLMs as the encoder, which we find excessive for DyTAG representation.


\subsection{Continuous-Time Dynamic Graph Learning}
Temporal graph neural networks (TGNNs) aim to process the dynamic graphs with/without vector-based edge/node attributes, which have been extensively studied due to their practical applicability. Existing TGNNs~\cite{kumar2019predicting,trivedi2019dyrep,DBLP:conf/wsdm/SankarWGZY20,xu2020inductive,tgn,wang2021inductive,ma2024temporal,ji2023community,DBLP:conf/aaai/LiYZC0ZTWM23,xu2024scalable,DBLP:conf/icde/000200O024,DBLP:conf/kdd/ZhongVYA24,DBLP:conf/kdd/0003MY024,wu2024feasibility,zhu2024topology,luimproving,bastosbeyond,gravina2024long,su2025temporal,qian2024mdgnn,yuan2025dg,li2025practicable,yin2024dynamic} typically focused on complex dynamic structure learning, where some~\cite{kumar2019predicting,tgn} leveraged sequence models (\textit{e.g.}, Gated Recurrent Unit (GRU)) to capture temporal dependencies. For example, co-neighbor encoding~\cite{yu2023towards,DBLP:conf/iclr/TianQG24,cheng2024co,zhang2024towards} and lightweight mixers~\cite{congwe,DBLP:conf/iclr/TianQG24,li2025ranking} were explored to model intricate structural dynamics, yet they treat timestamps as mere auxiliary, missing global temporal context and leaving temporal modality underexplored. Beyond dynamic structure modeling, significant efforts~\cite{wen2022trend,DBLP:conf/icml/GravinaLGBG24,DBLP:conf/iclr/SuZ024,tian2024latent,ma2024temporal} have been devoted to capturing long-range temporal dynamics through various techniques. However, the approaches mentioned above fall short in capturing the complex semantics of text attributes, as they often treat the textual modality as auxiliary or simply overlook them during structure learning, leading to sub-optimal performance for DyTAG representation.

\section{Preliminary}~\label{sec:3}
In this paper, we focus on dynamic text-attributed graphs and introduce the definition below.
\begin{definition}[Dynamic Text-Attributed Graphs (DyTAGs)]\label{def:1}
A Dynamic Text-Attributed Graph (DyTAG) can be defined as $\mathcal{G} = \{\mathcal{V}, \mathcal{E}\}$, where $\mathcal{V}$ represents the set of nodes, $\mathcal{E}\in \mathcal{V}\times \mathcal{V}$ denotes the set of edges. Let $\mathcal{T}$ denote the set of timestamps, $\mathcal{D}$ and $\mathcal{R}$ are the sets of node and edge text attributes, respectively. Each $v\in \mathcal{V}$ is associated with a text attribute $d_v\in \mathcal{D}$. Each $(u, v)\in \mathcal{E}$ can be represented as $(r_{uv}, t_{uv})$ with a text attribute $r_{uv}\in \mathcal{R}$ and a timestamp $t_{uv}\in \mathcal{T}$ to indicate the occurrence time of this edge. 
\end{definition}

Based on the definition~\ref{def:1}, DyTAGs can be regarded as continuous-time dynamic text-attributed graphs. We analyze the DyTAGs from the \textit{node-centric} perspective and highlight the characteristics of three modalities of real DyTAGs in Fig.~\ref{fig:1b} and Fig. 1 of the Appendix. Next, we follow an encoder-decoder framework~\cite{kazemi2020representation} and define the problem of representation learning on DyTAGs.


\begin{problem}[Representation Learning on Dynamic Text-Attributed Graphs]
    Given a DyTAG $\mathcal{G} = \{\mathcal{V}, \mathcal{E}\}$ with a sequence
of temporal events, \textit{e.g.}, $(u, v): (r_{uv}, t_{uv})$, we aim to
design an encoder function $F$: $(u, v)\rightarrow \mathbf{z}_u(t_{uv}), \mathbf{z}_v(t_{uv})\in \mathbb{R}^{f}$, where $\mathbf{z}_u(t_{uv})$ and $\mathbf{z}_v(t_{uv})$ respectively represents representations/embeddings of nodes $u$ and $v$ at time $t_{uv}$ and $f$ is the embedding dimension.
\end{problem}

Then, node representations are input into decoder functions for downstream tasks, such as link prediction. 



\section{The Proposed MoMent}

\subsection{Overview}
This paper proposes to model, integrate, and align multiple modalities for link prediction over DyTAGs. The key idea behind MoMent is to individually extract node correlations from three modalities and align them effectively to enhance node representations by a lightweight model design.
Concretely, we first construct multi-modal inputs from the node-centric view and then leverage self-attention-only encoders to encode temporal patterns, textual semantics, and structural correlations, which can extract unique and complementary features from DyTAGs. Then, we fuse these tokens from three encoders to generate node representations. To align modality-specific tokens, we design a dual-domain alignment loss that globally aligns temporal-textual distributions and then fine-tunes the multi-modal coherence at the instance level, which can avoid disjoint subspaces and ensure coherent node representations.

\subsection{Multi-Modal Modeling}
\subsubsection{Motivation}
As shown in Fig.~\ref{fig:1b}, modality-specific distributions are well separated, suggesting that each modality encodes distinct and complementary information. However, existing works~\cite{zhang2024dtgb,roy2025llm} typically treat timestamps and node attributes as auxiliary inputs, rather than exploiting them as fundamental temporal and semantic signals. As a result, the temporal and textual modalities are often underutilized. Additionally, our analysis of DyTAG datasets in the DTGB benchmark~\cite{zhang2024dtgb} reveals that only $3$ out of the $14$ node and edge text lengths exceed $128$ tokens (see Fig.~\ref{fig:1c}), raising questions about the necessity of using LLMs for textual correlation extraction. Notably, in some datasets, node attribute texts are longer than those of edge attributes, underscoring the rich semantics and importance of node attributes, which is often overlooked. These insights motivate us to explore lightweight designs for multi-modal learning on DyTAGs.

\subsubsection{Formulation}
We begin by constructing modality-specific inputs, \textit{i.e.}, node text attributes, temporal densities, and local structures. We then model node correlations based on these inputs for the node-centric modeling, employing a self-attention-based encoder for each modality to generate tokens, which are fused to generate node representations of DyTAGs. Given a batch of temporal edges before time $t$: $\mathcal{G}_b(t) = \{\mathcal{V}_b(t), \mathcal{E}_b(t)\}$ with raw node text attributes $\mathcal{D}_b(t)$, raw edge text attributes $\mathcal{R}_b(t)$ and observed timestamps $\mathcal{T}_b$ in $\mathcal{G}_b(t)$, we present multi-modal modeling and fusion.


\noindent\textbullet\ \textbf{Textual modality modeling.} 
For DyTAGs, textual information appears in both node and edge attributes. Since edge texts are inherently tied to topological structures, we focus on modeling node text attributes to extract semantic context for textual modality modeling, leaving edge text attributes for structure learning for integrity. To capture the underlying semantics, we process the raw node textual inputs $\mathcal{D}_b(t)$ using a pre-trained language model. These vectors are then encoded via a self-attention mechanism to generate modality-specific textual tokens. Formally, this process is defined as
\begin{equation}\label{eq:1}
  \mathbf{Z}^{x}_{b}(t) = \operatorname{SAM}_x(\mathbf{D}_b(t)), \mathbf{D}_b(t) = \operatorname{FFN}(\operatorname{PLM}(\mathcal{D}_b(t))),
\end{equation}
where $\operatorname{PLM}(\cdot)$ is the Bert-base-uncased model~\cite{kenton2019bert}, followed by DTGB for text attribute vectorization. $\operatorname{FFN}(\cdot)$ is the feedforward network, which is used to regularize the dimension; $\operatorname{SAM}_{*}(\cdot)$ is the self-attention mechanism~\cite{waswani2017attention}. $\mathbf{Z}^{x}_{b}(t)$ denotes textual tokens, which preserve rich semantic context for enhancing downstream tasks such as link prediction.


\noindent\textbullet\ \textbf{Temporal modality modeling.} For each node (\textit{e.g.}, source node), we know its behavior timestamps, where each timestamp is shared with a destination node for an edge. If we only model current timestamps for nodes, it is hard to provide discriminative information, as the source and destination nodes share the same timestamps. Thus, we attempt to model temporal densities for each node within the time gap to capture temporal patterns for each node, like activity cycles and frequency. Given $\mathcal{G}_b(t) = \{\mathcal{V}_b(t), \mathcal{E}_b(t)\}$ and observed timestamps $\mathcal{T}_b$ in $\mathcal{G}_b(t)$, we first extract their behavior timestamps within a time gap for each node and then input them into temporal encoder. We formulate the process as 

\begin{align}
\mathbf{T}_b(t) & =  \operatorname{FFN}([\phi(\mathcal{T}_{b})\Vert \phi(\mathcal{T}^{\iota})| \{t_{uv}^{\iota}\}_{v\in \mathcal{N}_{u}^{\iota}(t)}]),\label{eq:2}\\
\mathbf{Z}^{\tau}_{b}(t) &= \operatorname{SAM}_{\tau}(\operatorname{Pooling}(\mathbf{T}_b(t))),\label{eq:3}   
\end{align}
where $u\in \mathcal{V}_b(t)$ and $\mathcal{N}_{u}^\iota(t)$ denotes the node behaviors for node $u$ between $(t-\iota, t]$. $\phi(\cdot)$ is the time encoding function~\cite{xu2020inductive}. $\operatorname{Pooling}(\cdot)$ denotes the mean pooling function, $\Vert$ denotes concatenation, and zero-padding is applied to maintain uniform sequence length. $\mathbf{Z}^{\tau}_{b}(t)$ is the temporal tokens, which average the timestamps of the recent behaviors and encode recent temporal densities. Such temporal density and behavioral patterns serve as informative signals for distinguishing node states.

\noindent\textbullet\ \textbf{Structural modality modeling.} Unlike traditional structure learning methods, we focus on utilizing the local structures of nodes to capture their correlations to generate structural tokens for final node representations. Given $\mathcal{G}_b(t) = \{\mathcal{V}_b(t), \mathcal{E}_b(t)\}$, observed timestamps $\mathcal{T}_b$, and raw edge textual attributes $\mathcal{R}_b(t)$ in $\mathcal{G}_b(t)$, we first sample nodes' neighbors from previous timestamps and stack their features into sequences. These sequences are then fed into a structural encoder built on the self-attention mechanism to learn node correlations. We formulate the process as
\begin{align}
 \mathbf{S}_{b}(t) &= [\operatorname{PLM}(\mathcal{R}^{\varsigma}(t))\Vert\phi(\Delta\mathcal{T}^{\varsigma})|\{(r_{uv}^{\varsigma}, t_{uv}^{\varsigma})\}_{ v\in\mathcal{N}_{u}(t)}],\label{eq:3}\\   
 \mathbf{Z}^{s}_{b}(t) &= \operatorname{Pooling}(\operatorname{SAM}_{s}(\mathbf{S}_{b}(t))),\label{eq:4}
\end{align}
where $\Delta\mathcal{T}^{\varsigma}$ denotes the time gap between nodes and their neighbors, and $\mathcal{N}_{u}(t)$ represents the neighbors of node $u$ before timestamp $t$. Eq.~\eqref{eq:4} captures a node’s local neighbors at a given timestamp, preserving structural dependencies together with edge attributes. The resulting structural tokens, denoted as $\mathbf{Z}^{s}_{b}(t)$, encode node correlations along with these structural features. The structural tokens reflect external semantic dynamics, representing how a node interacts with and is generated based on its neighbor information. In contrast, node attributes and temporal density can be regarded as internal semantic dynamics, capturing the intrinsic characteristics and temporal evolution of the node itself. Last, we obtain three modality-specific tokens for each node.

\noindent\textbullet\ \textbf{Modality-specific token fuison.} Considering the difference between internal and external semantic dynamics, we design a dual fusion strategy, allowing the model to first aggregate internal semantic dynamics before incorporating structural context. Given learned temporal tokens $\mathbf{Z}^{\tau}_b(t)$, semantic tokens $\mathbf{Z}^{x}_b(t)$, and structural tokens $\mathbf{Z}^{s}_b(t)$, we generate the final node embeddings $\mathbf{Z}_b(t)$ by 
\begin{equation}\label{eq:6}
    \mathbf{Z}_b(t) = \mathbf{Z}^{s}_b(t) + \beta \mathbf{Z}^{\pi}_b(t), \; \mathbf{Z}^{\pi}_b(t) = \operatorname{FFN}(\mathbf{Z}^{x}_b(t) + \gamma \mathbf{Z}^{\tau}_b(t)),
\end{equation}
where $\beta$ and $\gamma$ are learnable parameters to adaptively balance their contribution.
 Eq.~\eqref{eq:6} allows the MoMent to jointly capture \textit{who the node is} and \textit{how it connects}, leading to comprehensive representations. In summary, our multi-modal network employs a node-centric modeling to leverage the complementary strengths of structural, temporal, and textual modalities. Unlike edge-centric modeling that performs early feature fusion before structure learning, our approach preserves modality-specific semantics through progressive integration. This design leads to more expressive representations. The following theorem provides theoretical support, showing that incorporating additional internal semantic dynamics guarantees increased task-relevant information.



\begin{theorem}[Conditional Information Gain]\label{theo:info_gain}
Let the multi-modal representation be $\mathbf{Z}(t)=\mathbf{Z}^{s}(t) + \beta\mathbf{Z}^{\pi}(t),$ and assume \(\beta \neq 0\). Then the following holds:
\begin{equation}
    I\!\bigl(\mathbf{Z}(t); \mathbf{Y}\bigr)
    = I\!\bigl(\mathbf{Z}^{s}(t); \mathbf{Y}\bigr) + I\!\bigl(\mathbf{Z}^{\pi}(t); \mathbf{Y}\mid \mathbf{Z}^{s}(t)\bigr),
\end{equation}
where $I(\cdot;\cdot)$ denotes mutual information.
\end{theorem}




\begin{proof}
By the chain rule of mutual information,
\begin{equation}\label{eq:P8}
    \begin{split}
       I\!\bigl(\mathbf{Z}(t); \mathbf{Y}\bigr)
&= I\!\bigl(\beta\mathbf{Z}^{\pi}(t)+\mathbf{Z}^{s}(t); \mathbf{Y}\bigr) \\[4pt]
&= I\!\bigl(\mathbf{Z}^{s}(t); \mathbf{Y}\bigr) \;+\;
   I\!\bigl(\beta\mathbf{Z}^{\pi}(t); \mathbf{Y}\mid \mathbf{Z}^{s}(t)\bigr). 
    \end{split}
\end{equation}
Since scaling $\beta$ by a non-zero value is a bijective transformation,
$I\!\bigl(\beta\mathbf{Z}^{\pi}(t); \mathbf{Y}\mid \mathbf{Z}^{s}(t)\bigr)
     = I\!\bigl(\mathbf{Z}^{\pi}(t); \mathbf{Y}\mid \mathbf{Z}^{s}(t)\bigr)$~\cite{cover1999elements}.
Substituting it into Eq.~\eqref{eq:P8} completes the proof.
\end{proof}

Theorem~\ref{theo:info_gain} indicates that incorporating the internal semantic dynamics $\mathbf{Z}^{\pi}(t)$ improves the information gain for predicting $\mathbf{Y}$, unless $I(\mathbf{Z}^{\pi}(t); \mathbf{Y} \mid \mathbf{Z}^s(t)) = 0$. In contrast, existing works~\cite{zhang2024dtgb,roy2025llm} fully or partially treat internal modalities as auxiliary information during structure learning, generating only structure-specific representations $\mathbf{Z}^s(t)$ and thereby missing the conditional gain $I(\mathbf{Z}^{\pi}(t); \mathbf{Y} \mid \mathbf{Z}^s(t))$ from internal temporal and semantic signals of nodes. This ensures that our multi-modal modeling not only preserves but enhances task-relevant signals, leading to higher-quality node representations and improved performance on downstream tasks.

\subsection{Dual-Domain Alignment Loss}
\subsubsection{Motivation} 
While MoMent can preserve the unique characteristics of each modality by multi-modal modeling, this may result in a disjoint latent space due to disjoint original distributions, compromising model robustness. Additionally, structural tokens are obtained by aggregating external semantic dynamics of nodes from their neighbors; in contrast, temporal and textual tokens are formed from internal semantic dynamics regardless of neighbor information. Given this distinction, we focus on aligning the internal semantic dynamics (temporal and textual modalities) first and then ensuring temporal-semantic-structural coherence.

\subsubsection{Formulation}
To achieve this, we propose a dual-domain alignment loss, which combines a distribution-level objective for global alignment and an instance-level objective for final alignment. Concretely, we first align tokens capturing internal semantic dynamics from temporal and textual modalities. As shown in Fig.~\ref{fig:1b}, these modalities exhibit non-overlapping peaks and different density distributions. After processing by non-shared encoders, the temporal tokens $\mathbf{Z}^{\tau}_b(t)$ and textual tokens $\mathbf{Z}^{x}_b(t)$ easily diverge. To mitigate this drift, we aim to reduce the discrepancy between their distributions using Maximum Mean Discrepancy (MMD), as it effectively captures complex statistical differences between audio-visual-language modalities~\cite{Li_2025_CVPR}. The distribution-level alignment loss is defined as
\begin{equation}\label{eq:7}
    \mathcal{L}_{\text{Distribution}} = \left\lVert \frac{1}{B}\sum_{i=1}^{B} \mathbf{z}^{\tau}_i(t)
      -\frac{1}{B}\sum_{j=1}^{B} \mathbf{z}^{x}_j(t) \right\rVert_{2}^{2},
\end{equation}
where $\left\lVert\cdot \right\rVert_{2}^{2}$ means the squared Euclidean norm and $B$ denotes the number of nodes in the batch. Eq.~\eqref{eq:7} establishes global alignment between the textual and temporal token distributions, but achieving comprehensive multi-modal representations also requires the integration of structural modality. Upon the dual token fusion in Eq.~\eqref{eq:6}, we design an instance-level alignment loss to synchronize further the fused temporal-textual tokens $\mathbf{Z}^{\pi}_b(t)$ with structural tokens $\mathbf{Z}^{s}_b(t)$. This alignment ensures cross-modal coherence by fine-tuning both tokens at the instance level. Specifically, we adopt cosine similarity to encourage semantic closeness between modalities, and define the alignment loss as
\begin{equation}\label{eq:8}
    \mathcal{L}_{\text{Instance}} = 1- \frac{1}{B}\sum_{i=1}^{B}
\frac{{\mathbf{z}^{\pi}_{i}(t)}^{\top}\mathbf{z}^{s}_i(t)}{\lVert \mathbf{z}^{\pi}_i(t)\rVert_2\,\lVert \mathbf{z}^{s}_i(t)\rVert_2},
\end{equation}
where $\left\lVert\cdot \right\rVert_{2}$ means the Euclidean norm.
Based on Eq.~\eqref{eq:8}, it fine-tunes the modality alignment, enforcing both internal coherence between temporal and semantic modalities and external coherence between node-centric and structure-aware representations. By mitigating the risk of disjoint feature subspaces across modalities, it enhances the coherence and quality of the final node representations.

The proposed alignment loss serves as an auxiliary loss that can be seamlessly integrated into the main loss function of link prediction. Thus, we define the overall loss as
\begin{equation}
   \mathcal{L} = \mathcal{L}_{\text{bce}} + \alpha\mathcal{L}_{\text{align}},\; \mathcal{L}_{\text{align}} =  \mathcal{L}_{\text{Distribution}} + \mathcal{L}_{\text{Instance}},
\end{equation}
where $\mathcal{L}_{\text{bce}}$ is Binary Cross Entropy (BCE) loss for link prediction. Other task losses for different downstream tasks can replace it.
MoMent preserves the main optimization objective while ensuring modality-specific tokens are aligned within a shared latent space. Thus, our MoMent can minimize modality discrepancies between three modality-specific tokens. Once aligned, decision-level fusion via Eq.~\eqref{eq:6} becomes more seamless, resulting in comprehensive and coherent node representations.

\begin{table*}[tbp]
	\caption{Performance on inductive link prediction under AUC and AP. The \colorbox{green!35}{\textbf{best}} and \colorbox{green!10}{\underline{second}} results are highlighted. `OOM' indicates an out-of-memory error, while `–' denotes unavailable results.}
	\label{tab:3}
	\centering
	\resizebox{\linewidth}{!}{
 \begin{tabular}{|c|c|c|c|c|c|c|c|}
        \Xhline{1pt}
        \rowcolor{gray!40} Approach & Enron & ICEWS1819 & Googlemap CT & Stack elec & Stack ubuntu & Amazon movies  & Yelp\\
        \hline
        \hline
        \multicolumn{8}{|c|}{\textbf{AUC under Inductive Setting}}\\
        \hline
        JODIE &0.8732 $\pm$ 0.0037&0.9285 $\pm$ 0.0065 &OOM &OOM &OOM &OOM &OOM \\
        DyRep &0.7901 $\pm$ 0.0047 & 0.9030 $\pm$ 0.0097 &OOM &OOM &OOM &OOM &OOM \\
        TGAT & 0.8650 $\pm$ 0.0032 &0.9706 $\pm$ 0.0054 &0.8791 $\pm$ 0.0028 &0.8423 $\pm$ 0.0018 &0.7655 $\pm$ 0.0019 &0.8706 $\pm$ 0.0023 &\cellcolor{green!10}\underline{0.9173 $\pm$ 0.0008} \\
        CAWN &0.9091 $\pm$ 0.0014 &\cellcolor{green!10}\underline{0.9774 $\pm$ 0.0039} &0.7058 $\pm$ 0.0047 &0.7963 $\pm$ 0.0074 &\cellcolor{green!10}\underline{0.7871 $\pm$ 0.0080} &0.8492 $\pm$ 0.0008 &0.8995 $\pm$ 0.0005 \\
        GraphMixer &0.8347 $\pm$ 0.0039&0.9605 $\pm$ 0.0025 & 0.7543 $\pm$ 0.0018 &0.8232 $\pm$ 0.0031 &0.7870 $\pm$ 0.0032 &0.8418 $\pm$ 0.0012 &0.8452 $\pm$ 0.0014 \\
        DyGFormer &0.9316 $\pm$ 0.0015&0.9630 $\pm$ 0.0027 & 0.7648 $\pm$ 0.0052& \cellcolor{green!10}\underline{0.8607 $\pm$ 0.0015} &0.7773 $\pm$ 0.0047 &\cellcolor{green!10}\underline{0.8733 $\pm$ 0.0005} &0.9067 $\pm$ 0.0009\\
        TPNet &0.8024 $\pm$ 0.0003 &0.9405 $\pm$ 0.0035 &0.7336 $\pm$ 0.0020 & 0.7877 $\pm$ 0.0242 &0.7564 $\pm$ 0.0010 &0.8184 $\pm$ 0.0009 &0.8122 $\pm$ 0.0014 \\
        \hline 
        LKD4DyTAG &\cellcolor{green!10}\underline{0.9367 $\pm$ 0.0011} &0.9543 $\pm$ 0.0013 & \cellcolor{green!10}\underline{0.8857 $\pm$ 0.2327} &- &- &- & -\\ 
        MoMent &\cellcolor{green!35}\textbf{0.9711 $\pm$ 0.0061} &\cellcolor{green!35}\textbf{0.9938 $\pm$ 0.0000}  &\cellcolor{green!35}\textbf{0.9560 $\pm$ 0.0031} &\cellcolor{green!35}\textbf{0.9482 $\pm$ 0.0066} &\cellcolor{green!35}\textbf{0.9599 $\pm$ 0.0012} &\cellcolor{green!35}\textbf{0.9837 $\pm$ 0.0171} &\cellcolor{green!35}\textbf{0.9713 $\pm$ 0.0102} \\
        \hline
        \hline
        \multicolumn{8}{|c|}{\textbf{AP under Inductive Setting}}\\
        \hline
        JODIE &0.8761 $\pm$ 0.0023 &0.9333 $\pm$ 0.0026 & OOM & OOM &OOM &OOM &OOM \\
        DyRep &0.7734 $\pm$ 0.0044&0.9134 $\pm$ 0.0041 &OOM & OOM &OOM &OOM &OOM\\
        TGAT &0.8589 $\pm$ 0.0031&\cellcolor{green!10}\underline{0.9716 $\pm$ 0.0033} & 0.8750 $\pm$ 0.0015 &0.8391 $\pm$ 0.0036 &0.7664 $\pm$ 0.0015 &0.8760 $\pm$ 0.0010 &\cellcolor{green!10}\underline{0.9174 $\pm$ 0.0010} \\ 
        CAWN &0.9223 $\pm$ 0.0011 &0.9631 $\pm$ 0.0034 &0.8012 $\pm$ 0.0021 &0.7921 $\pm$ 0.0086 &\cellcolor{green!10}\underline{0.7882 $\pm$ 0.0015 } &0.8508 $\pm$ 0.0006 &0.9010 $\pm$ 0.0009 \\
        GraphMixer &0.8328 $\pm$ 0.0034&0.9625 $\pm$ 0.0030 &0.7633 $\pm$ 0.0013 &0.8142 $\pm$ 0.0021 &0.7870 $\pm$ 0.0015 &0.8517 $\pm$ 0.0007 &0.8494 $\pm$ 0.0008\\
        DyGFormer &0.9409 $\pm$ 0.0025&0.9688 $\pm$ 0.0018&0.7735 $\pm$ 0.0031&\cellcolor{green!10}\underline{0.8801 $\pm$ 0.0043} &0.7832 $\pm$ 0.0015&\cellcolor{green!10}\underline{0.8780 $\pm$ 0.0006} &0.9092 $\pm$ 0.0006\\
        TPNet &0.7706 $\pm$ 0.0017 &0.9459 $\pm$ 0.0039 &0.7457 $\pm$ 0.0008 &0.7841 $\pm$ 0.0277 &0.7445 $\pm$ 0.0010 &0.8334 $\pm$ 0.0007 &0.8083 $\pm$ 0.0011\\ 
        \hline
        LKD4DyTAG &\cellcolor{green!10}\underline{0.9550 $\pm$ 0.0013} & 0.9634 $\pm$ 0.0010 &\cellcolor{green!10}\underline{0.8890 $\pm$ 0.1134} & - &- &- & - \\ 
        MoMent &\cellcolor{green!35}\textbf{0.9671 $\pm$ 0.0002} &\cellcolor{green!35}\textbf{0.9924 $\pm$ 0.0000} &\cellcolor{green!35}\textbf{0.9660 $\pm$ 0.0024} &\cellcolor{green!35}\textbf{0.9574 $\pm$ 0.0058} &\cellcolor{green!35}\textbf{0.9526 $\pm$ 0.0016} &\cellcolor{green!35}\textbf{0.9860 $\pm$ 0.0151} &\cellcolor{green!35}\textbf{0.9648 $\pm$ 0.0093}\\
       \Xhline{1pt}
\end{tabular}}
\end{table*}


\section{Experiments}

\subsection{Datasets, Baselines and Evaluation Metrics}
We collect $7$ datasets from~\cite{zhang2024dtgb}, which span four domains, such as the E-commerce network. We provide dataset statistics in Table 1 of the Appendix. To assess the effectiveness of MoMent, we compare it with eight baselines, including seven TGNNs based on the DTGB framework~\cite{zhang2024dtgb} and the only LLM-driven approach (LKD4DyTAG~\cite{roy2025llm}). We collect the results of JODIE~\cite{kumar2019predicting}, DyRep~\cite{trivedi2019dyrep}, TGAT~\cite{xu2020inductive}, CAWN~\cite{wang2021inductive}, GraphMixer~\cite{congwe}, and DyGFormer~\cite{yu2023towards} in Table~\ref{tab:3} from the DTGB benchmark~\cite{zhang2024dtgb}, and we run the DTGB with TPNet~\cite{luimproving} in our environments. We collect the results of LKD4DyTAG from its original paper.

We use link prediction as the downstream task in transductive and inductive settings. In the inductive setting, the test set consists of new nodes that do not appear in the training and validation sets. We use Average Precision (AP) and Area Under the Receiver Operating Characteristic Curve (AUC) as the metric~\cite{huang2024temporal,zhang2024dtgb}. We repeat all experiments five times and report both the mean and standard deviation of the results.

\noindent\underline{\textbf{The Appendix}} includes details on the \textit{datasets}, \textit{baselines}, \textit{training \& hyperparameter configurations}, \textit{additional experiments}, and \textit{source code}.

\subsection{Effectiveness and Efficiency Evaluation}\label{sec:5.2}
We evaluate the effectiveness of MoMent on link prediction under both transductive and inductive settings, using AUC and AP metrics. Due to space limits, results for the transductive setting are reported in Table 2 of the Appendix. We also evaluate the time efficiency and model generalization.

\noindent\textbf{Obs-1: MoMent outperforms TGNN-driven approaches due to additional information gain from internal semantic dynamics.}
As shown in Table~\ref{tab:3} and Table 2 of the Appendix, MoMent consistently outperforms seven TGNN-based baselines, achieving an average absolute improvement of $7.80\%$ in the inductive setting and $3.39\%$ in the transductive setting. This performance gain stems from MoMent’s ability to fully exploit the unique contributions of each modality in DyTAGs, rather than treating temporal and textual information as merely auxiliary inputs. Moreover, the proposed dual-domain alignment loss enforces coherence across temporal, textual, and structural modalities, leading to more coherent and task-relevant node representations. These empirical results align with our theoretical findings in Theorem~\ref{theo:info_gain}, which guarantees information gain through multi-modal modeling.

\noindent\textbf{Obs-2: MoMent outperforms the LLM-driven approach owing to comprehensive modality construction and modeling.} MoMent beats LKD4DyTAG in both transductive and inductive settings, with improvements of up to $7.03\%$ in AUC and $10.70\%$ in AP. This performance gap arises from LKD4DyTAG's edge-centric design, which relies heavily on LLMs and GNNs for edge representations while underutilizing internal features of nodes, such as temporal densities. In contrast, MoMent adopts a self-attention-only architecture. It leverages an effective feature construction to capture rich correlations across both internal semantic dynamics (temporal and textual features) and external semantic dynamics (neighbor information), leading to more comprehensive and coherent node representations.

\begin{figure}[t]
\centering
\subfloat[Training Time Cost]{
	\includegraphics[width=0.48\linewidth]{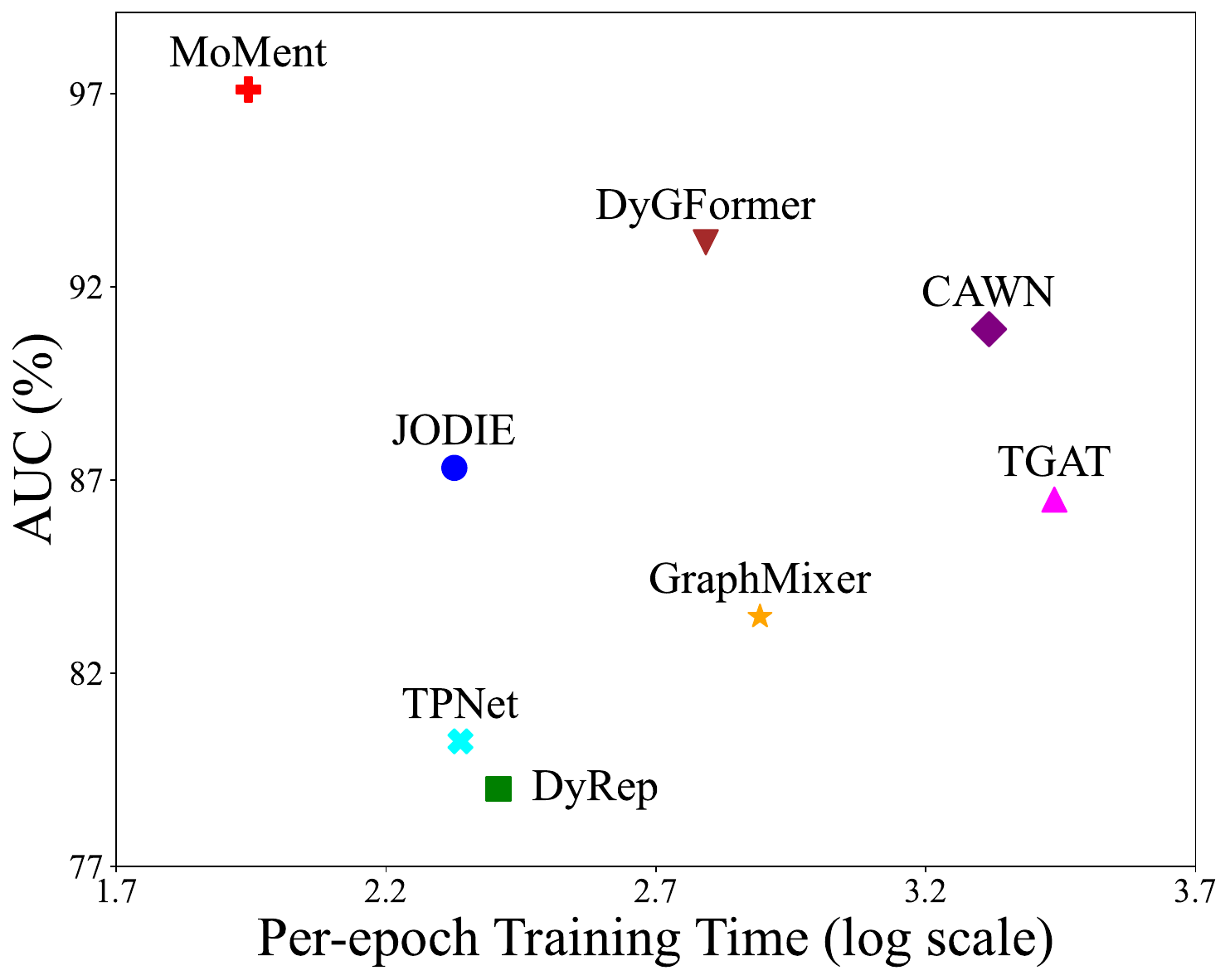}
		\label{fig:efficiency}
	}
\subfloat[Alignment Process]{
	\includegraphics[width=0.48\linewidth]{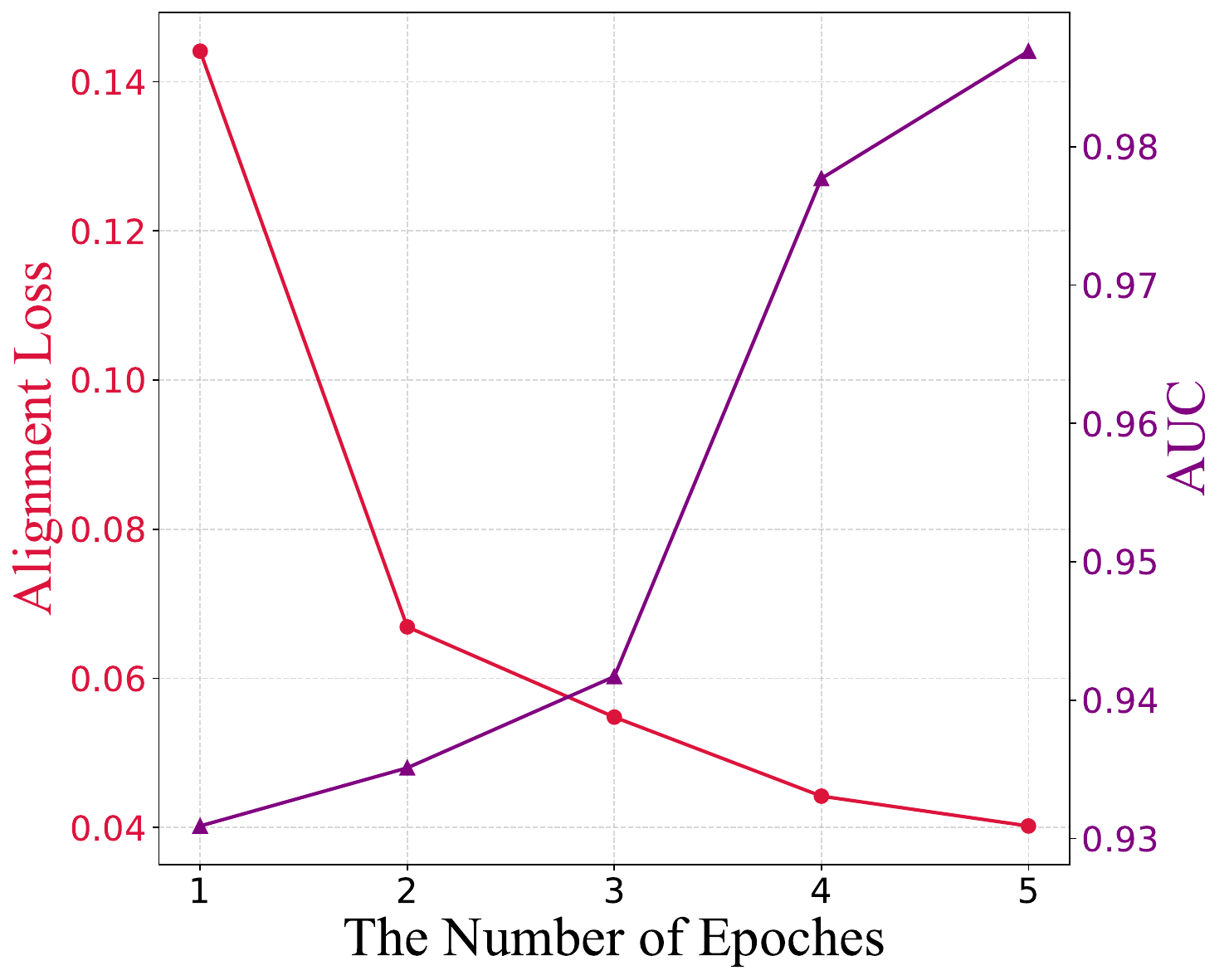}
		\label{fig:vensus}
	}
	\caption{(a) Resource consumption on Enron. (b) Averaged alignment loss value and dev AUC score on Enron across training epochs.} \label{fig:3}
\end{figure}

\noindent\textbf{Obs-3: MoMent achieves superior time efficiency due to lightweight architecture.} To assess time efficiency, we compare MoMent against seven TGNN baselines by measuring per-epoch training time on the Enron dataset~\footnote{The code of LKD4DyTAG is currently unavailable under the company policy, so its results are omitted in the efficiency evaluation.}. As shown in Fig.~\ref{fig:efficiency}, MoMent achieves $2\sim 31\times$ speed-up while showing better prediction accuracy on inductive link prediction. This is due to our multi-modal modeling with streamlined architecture without complex modules. In contrast, existing TGNNs emphasize structural modality, prioritizing the extraction of complex structural correlations, which results in substantial computational overhead.

\begin{table}[tbp]
	\caption{Ablation study under AUC in the inductive setting.}
	\label{tab:2}
	\centering
	\resizebox{\linewidth}{!}{
\begin{tabular}{|c|c|c|c|}
\Xhline{1pt}
\rowcolor{gray!40} Approach & Enron & Stack elec & Stack ubuntu \\
\hline
\hline
MoMent &\textbf{0.9711} &\textbf{0.9482} &\textbf{0.9599} \\
\hline
w/o Temporal.M &0.9427& 0.9059& 0.8534\\
w/o Textual.M &0.8387 & 0.7867 &0.7770 \\
w Structural Only &0.8671 & 0.7848 & 0.7660\\
w/o Align-D  & 0.9299 & 0.9294  & 0.9339 \\
w/o Align-I  &0.9482 & 0.9128 &0.9320 \\
w/o Align &0.9108 &0.9045 &0.9048 \\
\Xhline{1pt}
\end{tabular}}
\end{table}

\noindent\textbf{Obs-4: MoMent shows good generalization capabilities across different downstream tasks.} To assess the generalization capabilities of MoMent, we conduct edge classification experiments on three benchmark datasets, comparing it against eight state-of-the-art baselines. We report performance using weighted precision in Table 3 of the Appendix. MoMent consistently outperforms all baselines, achieving an average improvement of $2.64\%$ across datasets. This consistent gain highlights the effectiveness of our multi-modal network, which captures unique and complementary information from temporal, structural, and semantic contexts. This also confirms that MoMen's lightweight design allows it to generate high-quality node representations, leading to enhanced performance across various downstream tasks.


\subsection{Ablation Study}\label{sec:5.3}
We conduct six variants to assess the impact of individual modules using three datasets on inductive link prediction. Specifically, we evaluate: (1) removing the temporal modality modeling (`w/o Temporal.M'), (2) removing the textual modality modeling (`w/o Textual.M'), (3) using structural modality modeling only (`w Structural Only'), (4) removing the distribution-level alignment (`w/o Align-D'), (5) removing the instance-level alignment (`w/o Align-I'), and (6) removing the alignment loss (`w/o Align'). The AUC results in the inductive setting are reported in Table~\ref{tab:2}.

\noindent\textbf{Obs-5: Internal semantic dynamics complement external ones, where textual semantics contribute more significantly.} Modeling temporal and textual modalities offers significant discriminative power for link prediction, leading to average improvements of $5.91\%$ and $15.89\%$, respectively. This gain stems from unique characteristics in the original distributions of temporal patterns and textual semantics, which provide complementary information for structural modality. Node text semantics are critical for DyTAG representation, which is neglected by existing studies. In contrast, only external semantic dynamics result in a performance drop of up to $19.39\%$, highlighting the effectiveness of our multi-modal modeling and supporting Theorem~\ref{theo:info_gain}. 


\noindent\textbf{Obs-6: The dual-domain alignment loss facilitates learning coherent node representations.} Distribution-level and instance-level alignment contribute improvements of up to $4.12\%$ and $3.54\%$, respectively, on inductive link prediction, demonstrating the effectiveness of dual alignment across the three modalities. Moreover, removing the alignment loss leads to a performance drop of up to $6.03\%$, likely caused by disjoint latent subspaces of modality-specific tokens.

\subsection{Case Study and Visualization}
We further assess the effectiveness of modality alignment by analyzing its evolution during training, using case studies based on the Enron dataset. We visualize the modality-specific tokens at convergence on the Enron dataset.

\noindent\textbf{Obs-7: A positive correlation exists between modality alignment and prediction accuracy.} As shown in Fig.~\ref{fig:vensus}, this shows a clear inverse relationship between alignment loss value (red line) and development (dev) accuracy (purple line) across training epochs: as the alignment loss value decreases, AUC score increases. This suggests that better alignment between modality-specific tokens contributes directly to higher prediction accuracy, highlighting the significance of temporal-semantic-structural coherence in multi-modal modeling.

\begin{figure}[t]
\centering
\subfloat[Instance Alignment]{
	\includegraphics[width=0.48\linewidth]{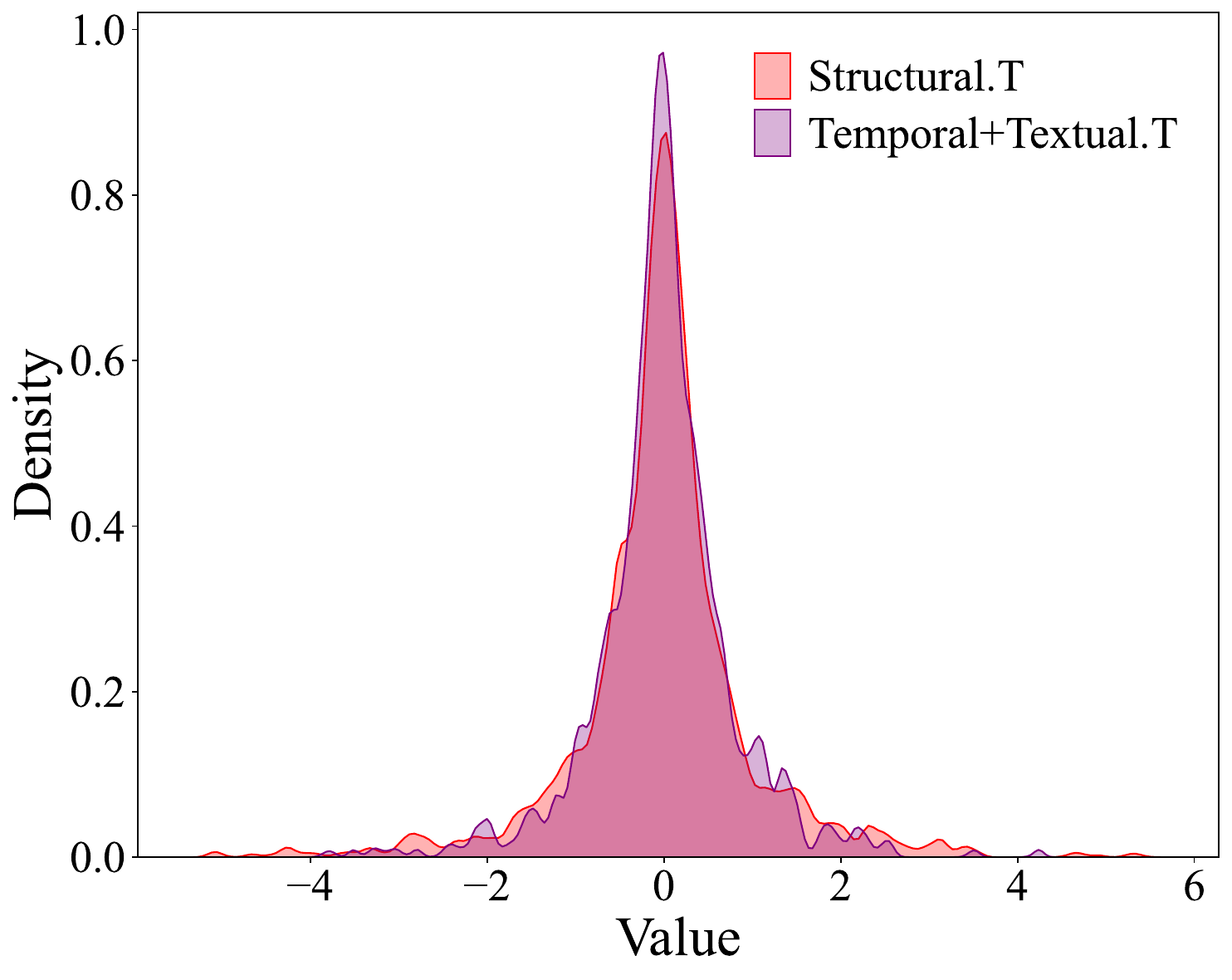}
		\label{fig:5a}
	}
\subfloat[Distribution Alignment]{
	\includegraphics[width=0.48\linewidth]{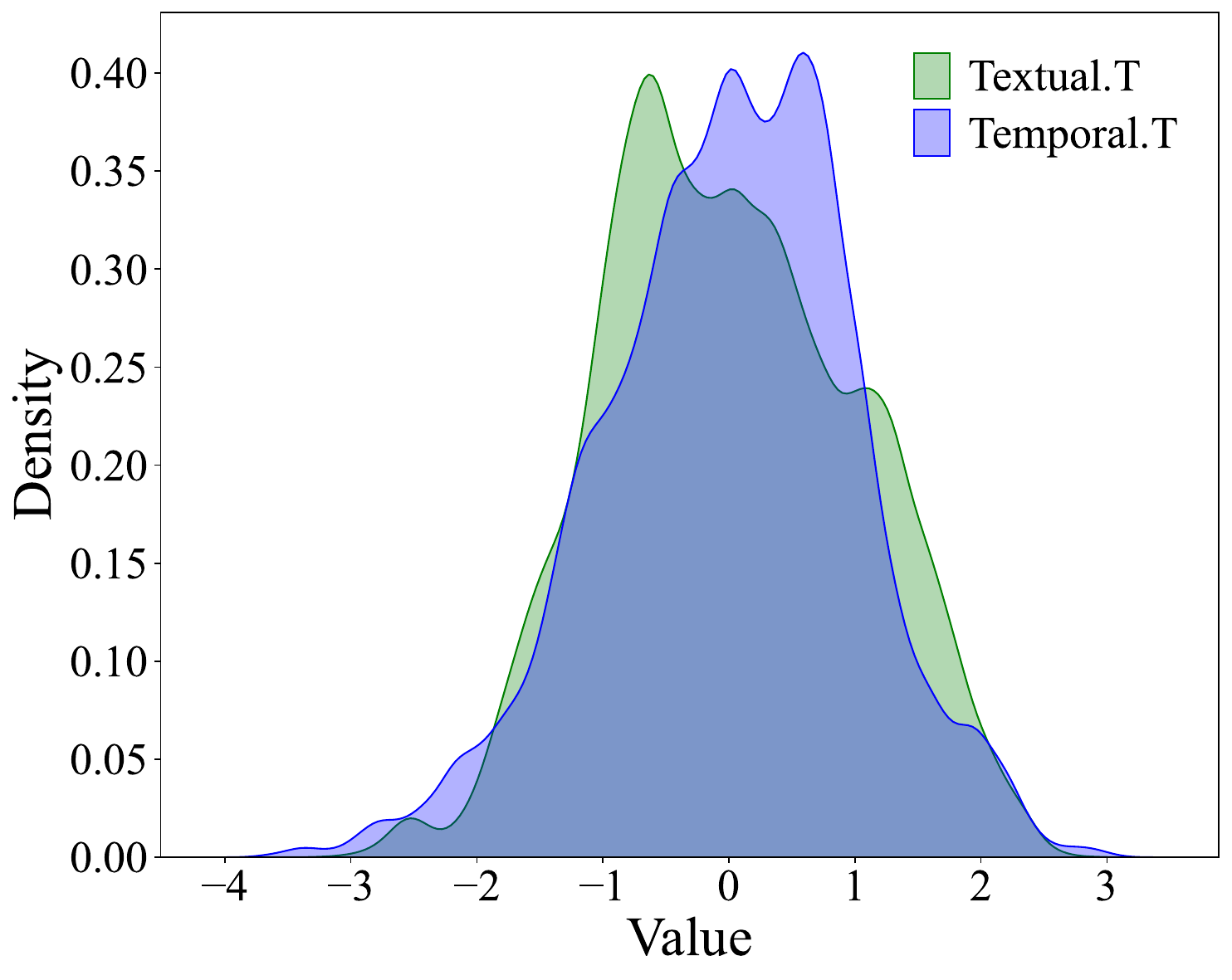}
		\label{fig:5b}
	}
	\caption{KDE distributions of tokens from three modalities on the Enron dataset. `X.T' indicates the use of learned tokens from each modality.} \label{fig:5}
\end{figure}

\noindent\textbf{Obs-8: MoMent shows coherent subspaces across multiple modalities.}
MoMent demonstrates effective modality alignment at both instance and distribution levels: (1) As shown in Fig.~\ref{fig:5a}, the distributions of structural and temporal+textual tokens exhibit significant overlap, indicating tight alignment and coherence across modalities at the instance level; and (2) Fig.~\ref{fig:5b} illustrates substantial overlap between temporal and textual token distributions, where slight differences in peak shapes suggest the preservation of modality-specific characteristics within a shared latent space. These observations confirm the effectiveness of our multi-modal modeling, alignment, and integration idea.

\section{Conclusion}
In this paper, we investigate the modeling, integration, and alignment of multi-modal features from a node-centric view for link prediction on dynamic text-attributed graphs. Unlike prior works that depend heavily on TGNNs and/or LLMs, we propose MoMent that captures correlations across textual semantics, temporal densities, and local structures by effective feature construction and encoding. The learned temporal, textual, and structural tokens are then integrated to generate node representations that preserve internal and external semantic dynamics. To enhance cross-modal coherence, we present a dual-domain alignment loss that globally aligns the distributions of temporal and textual tokens while enforcing coherence across three modalities at the instance level. We provide a theoretical analysis demonstrating that our multi-modal modeling offers greater information gain than TGNN-driven approaches. Extensive experiments confirm the effectiveness, efficiency, and generalization of our MoMent, and visualizations highlight its strength in multi-modal learning for DyTAGs.

\bibliography{aaai2026}

\appendix
\section{Additional Modality Visualization of DyTAGs}
\begin{figure*}[t]
    \subfloat[ICEWS1819]{\includegraphics[width=0.25\linewidth]{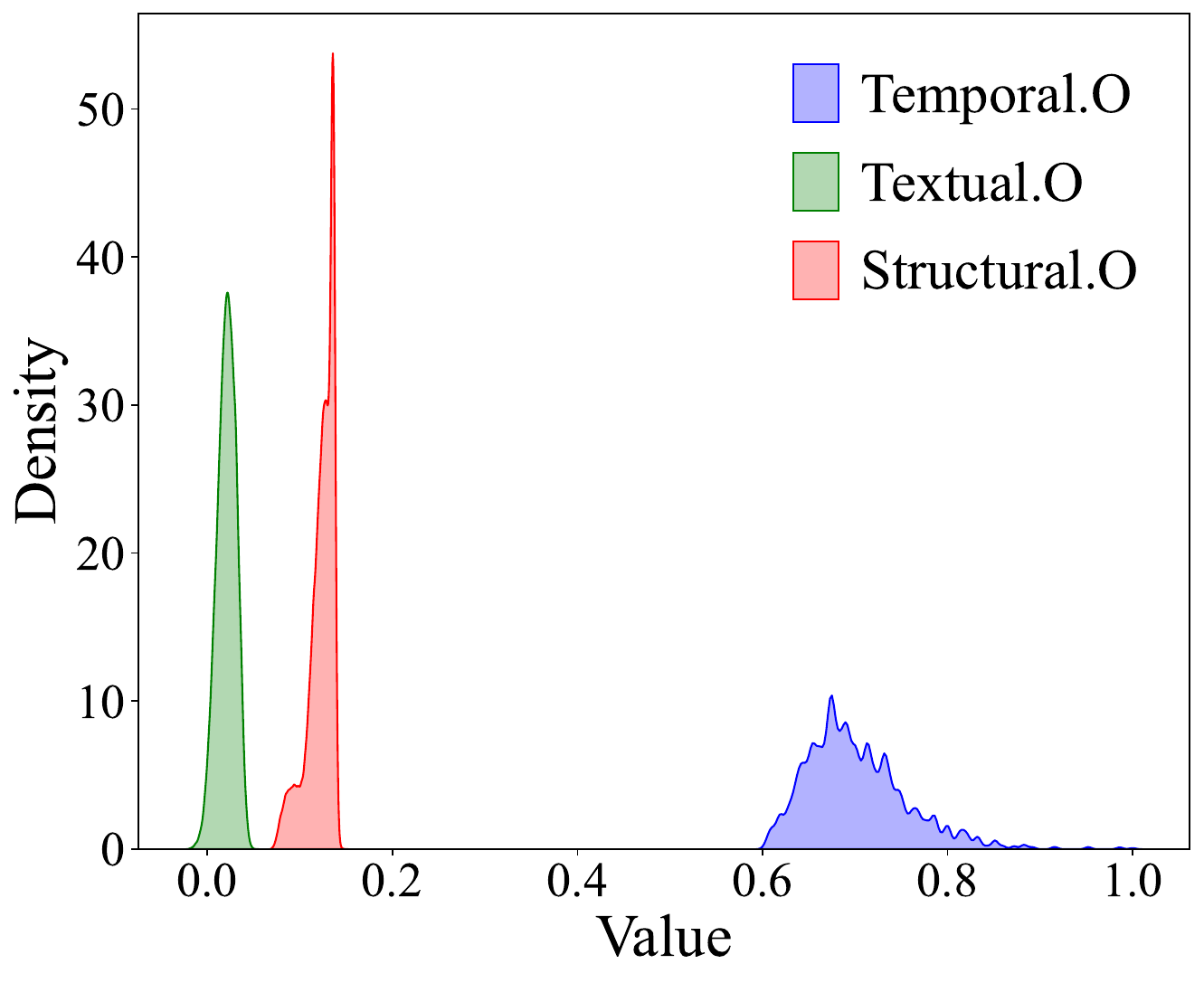}}
    \subfloat[Googlemap CT]{\includegraphics[width=0.25\linewidth]{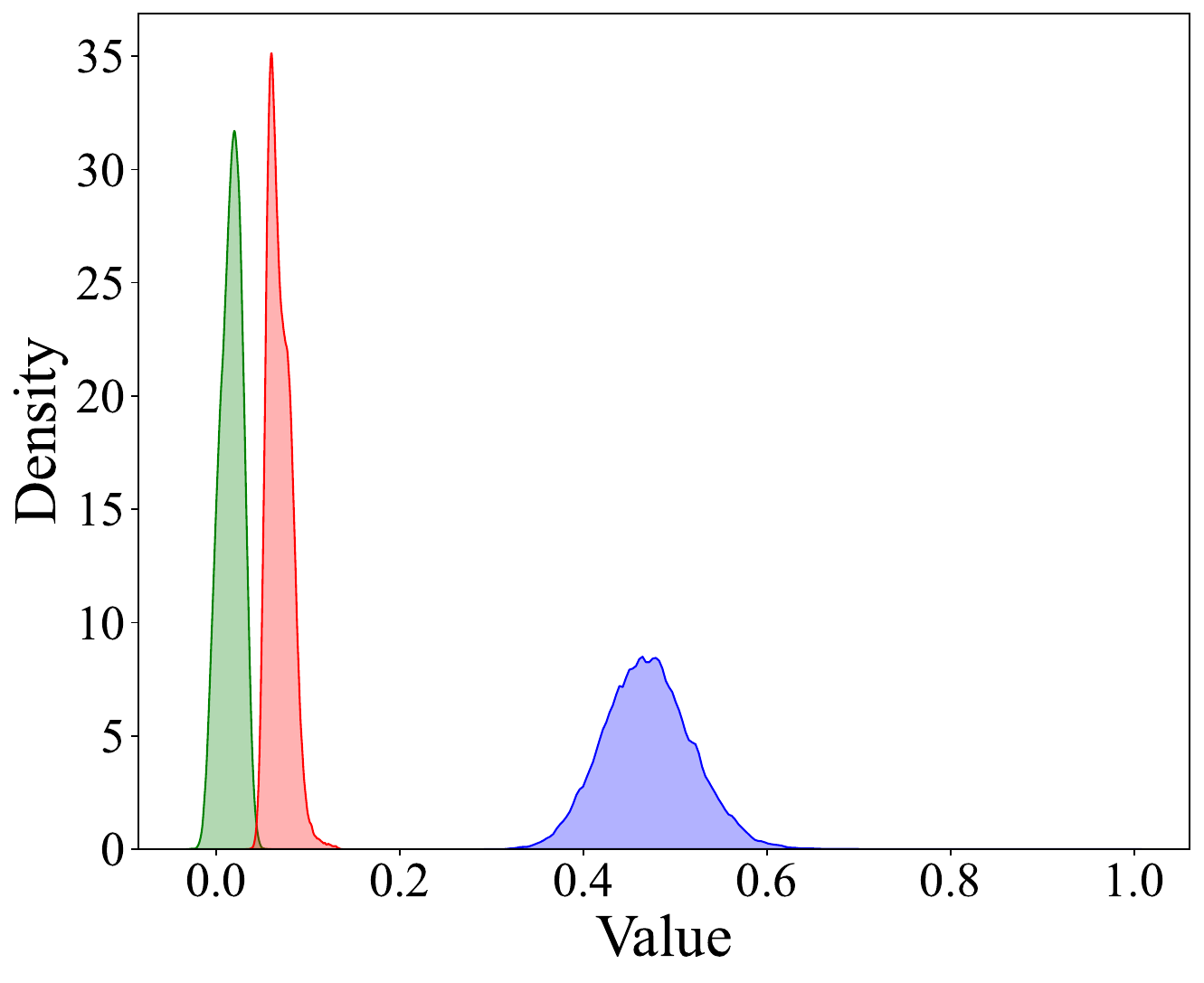}}
    \subfloat[Stack elec]{\includegraphics[width=0.25\linewidth]{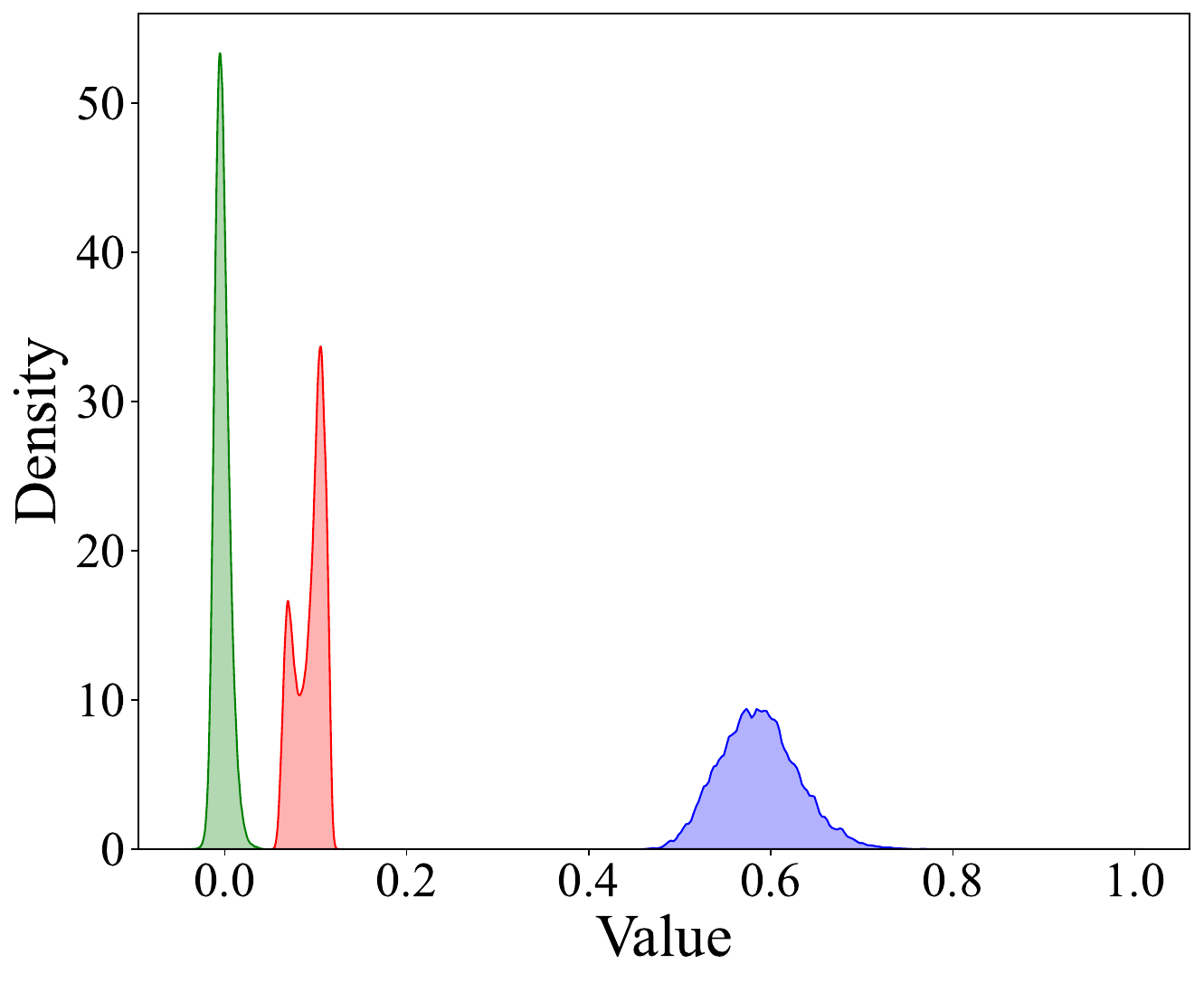}}
    \subfloat[Stack ubuntu]{\includegraphics[width=0.25\linewidth]{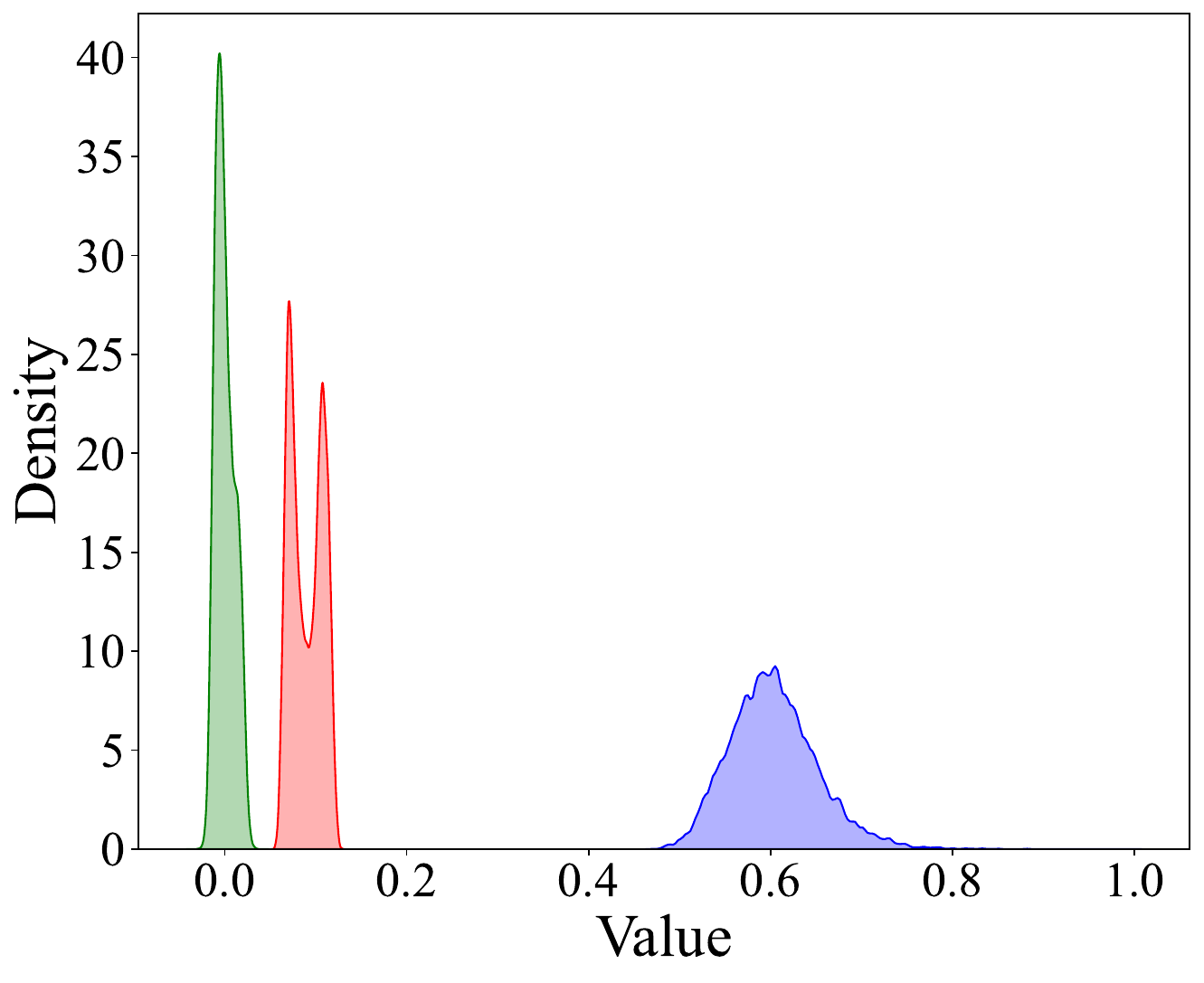}}
    \caption{KDE distribution of three modalities on four real dynamic text-attributed graphs, where `X.O' denotes the original features.}
    \label{fig:A7}
\end{figure*}

To further support our analysis of the multi-modal nature in DyTAGs, we provide additional KDE distribution visualizations across four datasets in Fig.~\ref{fig:A7}. Concretely, we take a \textit{node-centric} perspective to analyze the DyTAG and define its three modalities as
\begin{itemize}
    \item \underline{Structural modality} represents the dynamic graph topology, formed by evolving edges over time. This modality includes timestamps and textual attributes of edges, capturing dynamic structure changes as the graph continuously evolves.

    \item \underline{Temporal modality} is defined from a node-centric perspective. This modality represents the timestamps of node behaviors, reflecting intricate temporal evolutions within the graph.
    
    \item \underline{Textual modality} is defined from a node-centric perspective. This modality encapsulates the textual attributes of nodes, providing semantic information that complements structural and temporal aspects.
\end{itemize}



\section{Additional Experiments}

\subsection{Additional Dataset Description}\label{C:2}
We provide detailed statistics in Table~\ref{tab:1}, where we introduce the number of nodes, edges, edge categories, and timestamps as well as the type of text attributes. These datasets span various domains, including the email network (Enron), knowledge graph (ICEWS1819), multi-round dialogue (Stack elec and Stack ubuntu), and E-commerce network (Googlemap CT, Amazon movies, and Yelp). For more details of dataset preprocessing, please refer to the DTGB benchmark~\cite{zhang2024dtgb}.

\begin{table}[t]
  \caption{Dataset statistics.}
  \label{tab:1}
  \centering
  \small
  \resizebox{\linewidth}{!}{
  \begin{tabular}{|c|c|c|c|c|c|}
    \Xhline{1pt}
     \rowcolor{gray!40}Dataset & \# Nodes & \# Edges  &\#Categories & \# Timestamps & Text Attributes \\ 
    \hline
    \hline
    Enron &42,711 & 797,907 &10 & 1,006 & Node \& Edge\\
    ICEWS1819 & 31,796 & 1,100,071 & 266 & 730 & Node \& Edge\\
    Stack elec & 397,702 & 1,262,225&  2 & 5,224 & Node \& Edge\\
    Stack ubuntu & 674,248 & 1,497,006 & 2 & 4,972& Node \& Edge\\
    Googlemap CT &111,168 &1,380,623 &5 &55,521& Node \& Edge\\
    Amazon movies &293,566 &3,217,324 &5 &7,287 & Node \& Edge\\
    Yelp &2,138,242 &6,990,189 &5 &6,036 & Node \& Edge\\
    \hline
    \end{tabular}}
\end{table}

\begin{table*}[tbp]
	\caption{Performance on transductive link prediction under AUC and AP. The \colorbox{green!35}{\textbf{best}} and \colorbox{green!10}{\underline{second}} results are highlighted. `OOM' indicates an out-of-memory error, while `–' denotes unavailable results.}
	\label{tab:trans}
	\centering
	\resizebox{\linewidth}{!}{
 \begin{tabular}{|c|c|c|c|c|c|c|c|}
        \Xhline{1pt}
        \rowcolor{gray!40} Approach & Enron & ICEWS1819 & Googlemap CT & Stack elec & Stack ubuntu & Amazon movies  & Yelp\\
        \hline
        \hline
        \multicolumn{8}{|c|}{\textbf{AUC under Transductive Setting}}\\
        \hline
        JODIE & 0.9731 $\pm$ 0.0052 & 0.9741 $\pm$ 0.0113 &OOM &OOM &OOM &OOM &OOM \\
        DyRep &0.9274 $\pm$ 0.0026 &0.9632 $\pm$ 0.0027 &OOM &OOM &OOM &OOM &OOM \\
        TGAT &0.9681 $\pm$ 0.0026& 0.9904 $\pm$ 0.0039 &0.9049 $\pm$ 0.0071 &0.9709 $\pm$ 0.0014 &0.9490 $\pm$ 0.0018 &0.9064 $\pm$ 0.0014 &\cellcolor{green!10}\underline{0.9487 $\pm$ 0.0029} \\
        CAWN &0.9740 $\pm$ 0.0007 & 0.9857 $\pm$ 0.0018 &0.8687 $\pm$ 0.0063 & 0.9631 $\pm$ 0.0007 &\cellcolor{green!10}\underline{0.9567 $\pm$ 0.0004} & 0.8936 $\pm$ 0.0012 &0.9349 $\pm$ 0.0026 \\
        GraphMixer &0.9567 $\pm$ 0.0013 & 0.9863 $\pm$ 0.0024  &0.8095 $\pm$ 0.0014 &0.9673 $\pm$ 0.0011 &0.9494 $\pm$ 0.0028 &0.8894 $\pm$ 0.0008 &0.8927 $\pm$ 0.0021\\
        DyGFormer &0.9779 $\pm$ 0.0014 &0.9888 $\pm$ 0.0015&0.8207 $\pm$ 0.0018 & \cellcolor{green!10}\underline{0.9798 $\pm$ 0.0006} &0.9526 $\pm$ 0.0035 &\cellcolor{green!10}\underline{0.9100 $\pm$ 0.0006} &0.9407 $\pm$ 0.0010 \\
        TPNet &0.9475 $\pm$ 0.0010 & 0.9818 $\pm$ 0.0009 & 0.7960 $\pm$ 0.0014  & 0.9611 $\pm$ 0.0036  &0.9516 $\pm$ 0.0004 &0.8726 $\pm$ 0.0003 &0.8691 $\pm$ 0.0007\\
        \hline 
        LKD4DyTAG &\cellcolor{green!10}\underline{0.9887 $\pm$ 0.0011} &\cellcolor{green!10}\underline{0.9950 $\pm$ 0.0014} & \cellcolor{green!10}\underline{0.9127 $\pm$ 0.0037} &- &- &- & -\\ 
        MoMent &\cellcolor{green!35}\textbf{ 0.9906 $\pm$ 0.0013 } &\cellcolor{green!35}\textbf{ 0.9992 $\pm$ 0.0000}  &\cellcolor{green!35}\textbf{0.9797 $\pm$ 0.0009 } &\cellcolor{green!35}\textbf{ 0.9808 $\pm$ 0.0009 } &\cellcolor{green!35}\textbf{0.9810 $\pm$ 0.0094 } &\cellcolor{green!35}\textbf{ 0.9868 $\pm$ 0.0178 } &\cellcolor{green!35}\textbf{ 0.9819 $\pm$ 0.0001} \\
        \hline
        \hline
        \multicolumn{8}{|c|}{\textbf{AP under Transductive Setting}}\\
        \hline
        JODIE &0.9553 $\pm$ 0.0051 &0.9752 $\pm$ 0.0037 & OOM & OOM &OOM &OOM &OOM \\
        DyRep &0.9066 $\pm$ 0.0076 &0.9676 $\pm$ 0.0026 &OOM & OOM &OOM &OOM &OOM\\
        TGAT &0.9668 $\pm$ 0.0026 &0.9908 $\pm$ 0.0032 &0.9002 $\pm$ 0.0019 & 0.9646 $\pm$ 0.0005 &0.9040 $\pm$ 0.0056 &0.9065 $\pm$ 0.0016 &\cellcolor{green!10}\underline{0.9457 $\pm$ 0.0025}\\
        CAWN &0.9756 $\pm$ 0.0008 & 0.9886 $\pm$ 0.0025 &0.8721 $\pm$ 0.0027 & 0.9529 $\pm$ 0.0023 &0.9420 $\pm$ 0.0005 &0.8885 $\pm$ 0.0007 &0.9323 $\pm$ 0.0027 \\
        GraphMixer &0.9559 $\pm$ 0.0027 & 0.9871 $\pm$ 0.0034& 0.8072 $\pm$ 0.0010 & 0.9591 $\pm$ 0.0009 &0.9416 $\pm$ 0.0047 &0.8906 $\pm$ 0.0006 &0.8883 $\pm$ 0.0026\\
        DyGFormer &0.9804 $\pm$ 0.0015 &0.9901 $\pm$ 0.0018&0.8183 $\pm$ 0.0038  & \cellcolor{green!35}\textbf{0.9819 $\pm$ 0.0010} &\cellcolor{green!10}\underline{0.9431 $\pm$ 0.0008} &\cellcolor{green!10}\underline{0.9097 $\pm$ 0.0003} &0.9391 $\pm$ 0.0011\\
        TPNet &0.9373 $\pm$ 0.0013 & 0.9823 $\pm$ 0.0009 & 0.7944 $\pm$ 0.0013  & 0.9339 $\pm$ 0.0078  &0.9331 $\pm$ 0.0006 &0.8767 $\pm$ 0.0002 &0.8547 $\pm$ 0.0006\\ 
        \hline
        LKD4DyTAG &\cellcolor{green!35}\textbf{0.9943 $\pm$ 0.0001} &\cellcolor{green!10}\underline{0.9970 $\pm$ 0.0012} & \cellcolor{green!10}\underline{0.9144 $\pm$ 0.0013} & - &- &- & - \\ 
        MoMent & \cellcolor{green!10}\underline{0.9894 $\pm$ 0.0015}  &\cellcolor{green!35}\textbf{0.9991 $\pm$ 0.0000 }  &\cellcolor{green!35}\textbf{0.9824 $\pm$ 0.0004 } &\cellcolor{green!10}\underline{0.9730 $\pm$ 0.0033} &\cellcolor{green!35}\textbf{0.9729 $\pm$ 0.0125 } &\cellcolor{green!35}\textbf{ 0.9864 $\pm$ 0.0080 } &\cellcolor{green!35}\textbf{0.9896 $\pm$ 0.0002} \\
       \Xhline{1pt}
\end{tabular}}
\end{table*}

\subsection{Baseline}
We compare our MoMent against seven T-GNNs and one LLM-based method for the link prediction task. We introduce them below.
\begin{itemize}
    \item \underline{JODIE} leverages recurrent neural networks to update node memories and then map node memories into future representation trajectories of nodes.
    
    \item \underline{DyRep} regards each temporal event as a temporal point process and constructs node memory update and evolving structure aggregation for temporal representation.
    
    \item \underline{TGAT} designs a time encoding function and applies the self-attention mechanism for temporal and structural modeling to generate node representations.
    
    \item \underline{CAWN} designs causal anonymous walks to capture structure evolution and employs an attention mechanism to learn these sampled walks for downstream tasks.

    \item \underline{GraphMixer} attempts to design a simple network architecture for temporal graph learning and leverage the MLP Mixer~\cite{tolstikhin2021mlp} based on historical neighbors for node representations.
    
    \item \underline{DyGFormer} proposes a neighbor co-occurrence encoding scheme to capture neighbor correlations and integrates these encodings into a transformer architecture~\cite{vaswani2017attention} to generate node representations.

    \item \underline{TPNet} unifies existing relative encodings as a function of temporal walk matrices and efficiently encodes pairwise information through temporal walk matrix projection. It then leverages the MLP Mixer to generate node representations.
    
    \item \underline{LKD4DyTAG} employs knowledge distillation that transfers insights from LLM-based text edge representations to GNN-based spatio-temporal edge representations for different downstream tasks.

\end{itemize}

\subsection{Training Configuration} 
We run all experiments on a single machine with Intel(R) Core(TM) i9-10980XE 3.00GHz CPUs, NVIDIA RTX A6000, and $48$ GB RAM memory. We train the MoMent with Adam optimizer~\cite{kingma2014adam}, with an empirical learning rate of $0.00001$. Following the baseline~\cite{zhang2024dtgb}, we chronologically split each dataset into train/validation/test sets using a $7:1.5:1.5$ ratio.

\subsection{Hyperparameter Configuration}
Here, we present the hyperparameter configurations of our MoMent. These configurations remain unchanged across two downstream tasks and all datasets in our experiments. The details are provided as follows.

\begin{itemize}
    \item Dimension of embeddings/tokens: $768$ for the node representation dimension; $768$ for the structural token dimension; $128$ for the temporal token dimension; $128$ for the textual token dimension.
    \item Configurations for modality-specific encoders: $2$ for attention heads; $1$ for network layers; $512$ for hidden dimension; $0.1$ for dropout rate.
    \item Configurations for loss: $0.2$ for the hyperparameter $\alpha$.
\end{itemize}

Please refer to the default settings in DTGB~\cite{zhang2024dtgb}, TPNet~\cite{luimproving}, and LKD4DyTAG~\cite{roy2025llm} for the hyperparameter configurations of eight baselines.

\subsection{Additional Main Experimental Results}
We report additional experimental results of transductive link prediction and edge classification in Tables~\ref{tab:trans} and~\ref{tab:edge}, respectively. Statistical significance between comparative approaches is assessed with a Wilcoxon signed-rank test, where $\alpha = 0.05$. Overall Result analysis is provided in Section 5.2 in the main paper.

\begin{table}[tbp]
	\caption{Performance on edge classification under precision. The \colorbox{green!35}{\textbf{best}} and \colorbox{green!10}{\underline{second}} results are highlighted. `OOM' indicates an out-of-memory error.}
	\label{tab:edge}
	\centering
	\resizebox{\linewidth}{!}{
\begin{tabular}{|c|c|c|c|}
\Xhline{1pt}
\rowcolor{gray!40} Approach & Enron & Stack elec & Stack ubuntu \\
\hline
\hline
JODIE &0.6568 $\pm$ 0.0043&OOM&OOM\\
DyRep &0.6635 $\pm$ 0.0052 &OOM&OOM\\
TGAT &0.6148 $\pm$ 0.0012 &0.6265 $\pm$ 0.0046 & 0.6858 $\pm$ 0.0047 \\
CAWN &0.6076 $\pm$ 0.0070 &0.6167 $\pm$ 0.0094 &0.6921 $\pm$ 0.0040 \\
GraphMixer&0.6313 $\pm$ 0.0024 &0.6074 $\pm$ 0.0039 &0.6930 $\pm$ 0.0028\\
DyGFormer&0.6601 $\pm$ 0.0067 & 0.6026 $\pm$ 0.0471& 0.6789 $\pm$ 0.0490\\ 
TPNet&0.6488 $\pm$ 0.0045& \cellcolor{green!10}\underline{0.6993 $\pm$ 0.0038} & \cellcolor{green!10}\underline{0.7247 $\pm$ 0.0065}\\
\hline 
LKD4DyTAG& \cellcolor{green!10}\underline{0.6685 $\pm$ 0.0013} & 0.6565 $\pm$ 0.0056 & 0.6824 $\pm$ 0.0054\\
MoMent &\cellcolor{green!35}\textbf{0.6958 $\pm$ 0.0012} &\cellcolor{green!35}\textbf{0.7151 $\pm$ 0.0043} &\cellcolor{green!35}\textbf{0.7362 $\pm$ 0.0041} \\
\Xhline{1pt}
\end{tabular}}
\end{table}

\begin{figure}
    \centering
    \includegraphics[width=0.7\linewidth]{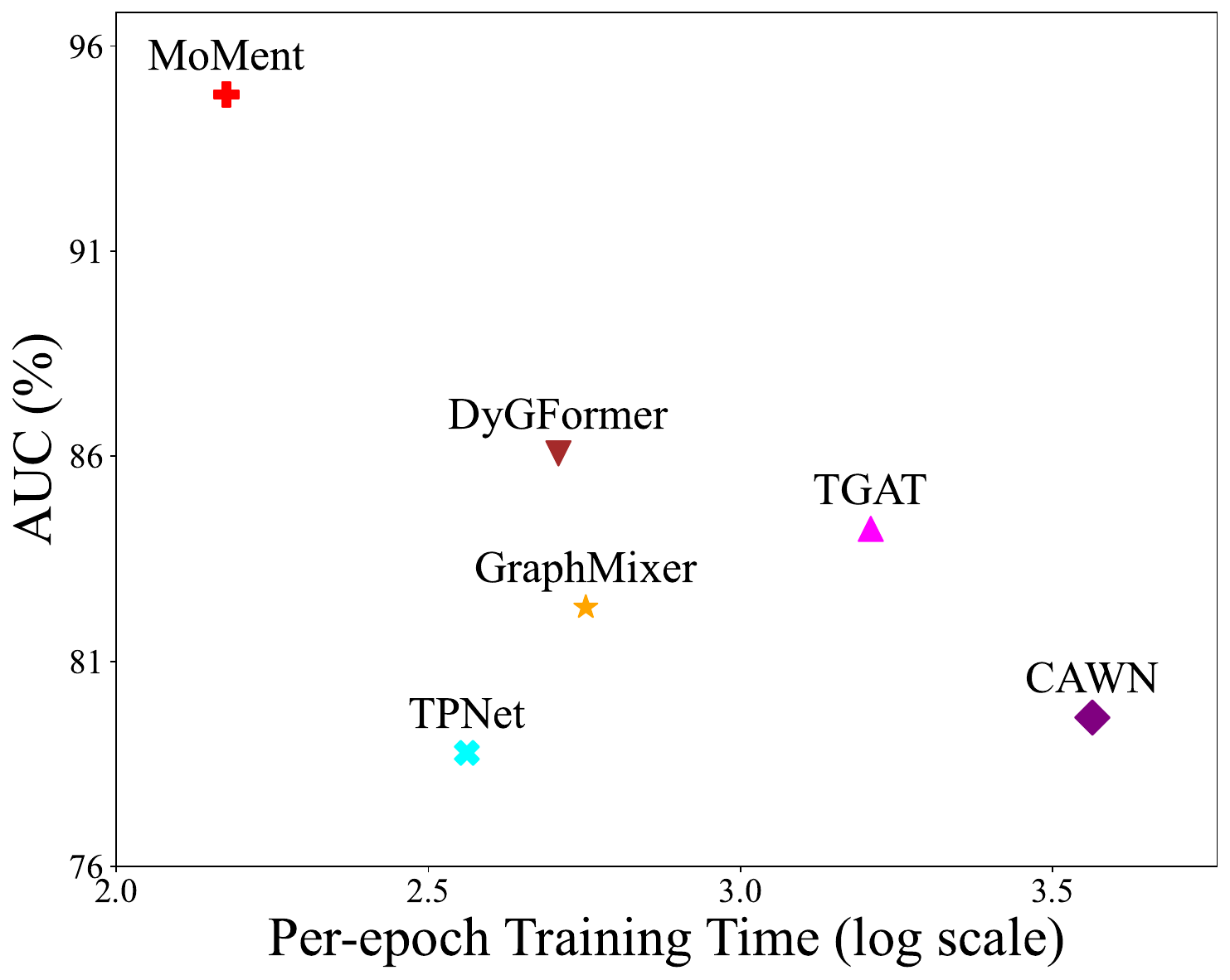}
    \caption{Resource consumption on the Stack elec dataset.}
    \label{fig:time}
\end{figure}

We also provide additional training time costs on the Stack elec dataset in Fig.~\ref{fig:time}. Our MoMent is faster than seven TGNN baselines. Note that both JODIE and DyRep ran out of memory.

\begin{figure}[h]
    \centering
    \includegraphics[width=0.7\linewidth]{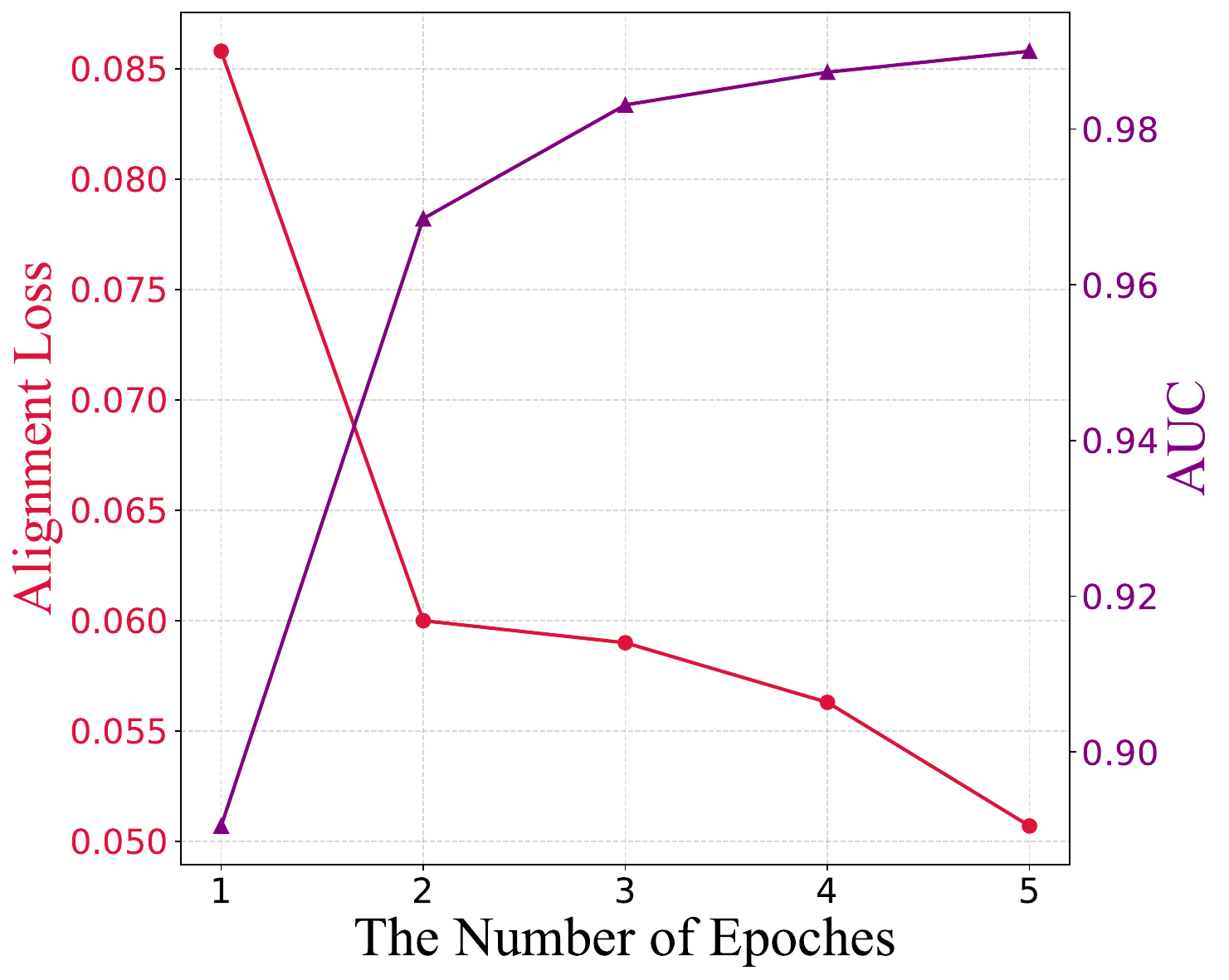}
    \caption{Averaged alignment loss value and dev AUC score on the Stack elec dataset across training epochs.}
    \label{fig:align}
\end{figure}

\subsection{Additional Case Study}
We show the evolution of MoMent during training by showing the relationships between modality alignment and development accuracy using the Stack elec dataset.
As shown in Fig.~\ref{fig:align}, this also shows a clear inverse relationship between alignment loss value (red line) and development (dev) accuracy (purple line) across training epochs, which confirms the effectiveness of our alignment loss.

\begin{figure}[t]
    \centering
    \subfloat[Stack elec]{\includegraphics[width=0.5\linewidth]{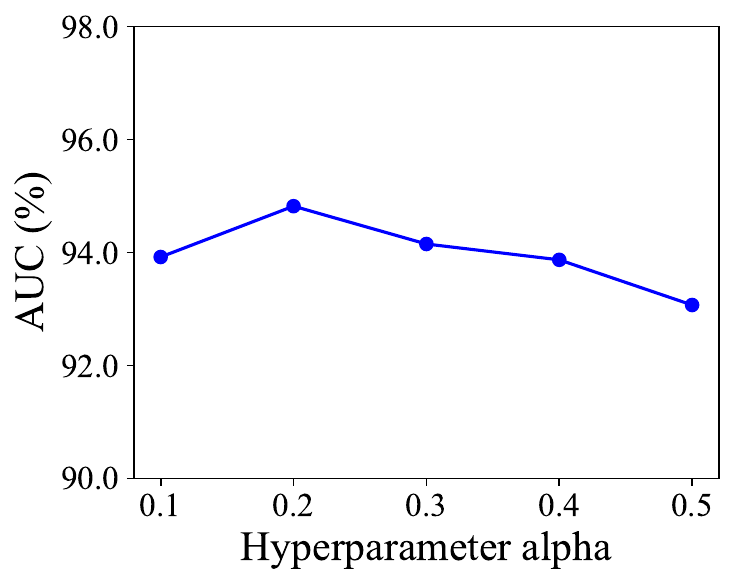}}
    \subfloat[Stack ubuntu]{\includegraphics[width=0.5\linewidth]{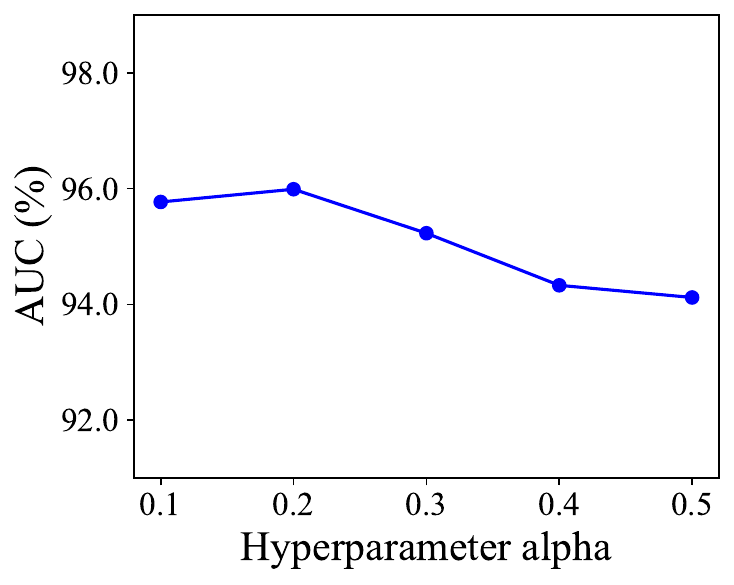}}
    \caption{Impact of the hyperparameter $\alpha$.}
    \label{fig:parameter}
\end{figure}

\subsection{Parameter Sensitivity Analysis }
We conduct experiments to evaluate the impacts of hyperparameter $\alpha$ in our MoMent, which is used for balancing the main loss with our alignment loss. We vary the $\alpha$ among $\{0.1, 0.2, 0.3, 0.4, 0.5\}$ and report AUC results on the inductive link prediction on two datasets (\textit{i.e.}, Stack elec and Stack ubuntu) in Fig.~\ref{fig:parameter}. 

\noindent\textbf{Obs-9: MoMent is relatively robust to hyperparameter changes.} Increasing the weight of the alignment loss improves model performance by emphasizing modality coherence. Additionally, MoMent remains robust to $\alpha$ when set as $0.1$ and $0.3$. However, when $\alpha$ exceeds $0.3$, it overwhelms the primary loss term and degrades overall performance.
Throughout the experiments, we set $\alpha$ as $0.2$.

\end{document}